\documentclass[twoside]{article}

\usepackage[accepted]{aistats2021}
\usepackage[round]{natbib}

\usepackage{wrapfig}
\usepackage{multirow}
\usepackage{amsmath}
\usepackage{mathtools}
\usepackage{amsfonts}       
\usepackage[ruled]{algorithm}
\usepackage{algpseudocode}
\usepackage{microtype}      
\usepackage{booktabs}       
\usepackage{makecell}
\usepackage{subfigure}
\usepackage{color}
\usepackage{IEEEtrantools}

\newcommand{\argmax}{\mathop{\rm argmax}}
\newcommand{\argmin}{\mathop{\rm argmin}}

\newcommand{\balpha}{\ensuremath{{\boldsymbol \alpha}}}
\newcommand{\bbeta}{\ensuremath{{\boldsymbol \beta}}}
\newcommand{\bgamma}{\ensuremath{{\boldsymbol \gamma}}}
\newcommand{\btheta}{\ensuremath{{\boldsymbol \theta}}}

\newcommand{\vio}[1]{{\color{black} #1}}

\newtheorem{proposition}{Proposition}

\begin{document}

%
\runningtitle{SDF-Bayes: Cautious Optimism in  Safe Dose-Finding Clinical Trials}

%
\runningauthor{Hyun-Suk Lee, Cong Shen, William Zame, Jang-Won Lee, Mihaela van der Schaar}

\twocolumn[

\aistatstitle{SDF-Bayes: Cautious Optimism in  Safe Dose-Finding Clinical Trials with Drug Combinations and Heterogeneous Patient Groups}

\vspace{-0.15in}
\aistatsauthor{ Hyun-Suk Lee \And Cong Shen \And  William Zame }

\aistatsaddress{ Sejong University \And  University of Virginia \And UCLA }
\vspace{-0.2in}
\aistatsauthor{ Jang-Won Lee \And Mihaela van der Schaar }

\aistatsaddress{  Yonsei University \And University of Cambridge, UCLA,\\The Alan Turing Institute }

\vspace{-0.15in}
]

\begin{abstract}
\vspace{-0.1in}
Phase I clinical trials are designed to test the safety (non-toxicity) of drugs and find the maximum tolerated dose (MTD). This task becomes significantly more challenging when multiple-drug dose-combinations (DC) are involved, due to the inherent conflict between the {\em exponentially} increasing DC candidates and the limited patient budget. This paper proposes a novel Bayesian design, {\em SDF-Bayes}, for finding the MTD for drug combinations in the presence of safety constraints.  Rather than the conventional principle of escalating or de-escalating the current dose of one drug (perhaps alternating between drugs), SDF-Bayes proceeds by {\em cautious optimism}: it chooses the next DC that, on the basis of current information, is most likely to be the MTD (optimism), subject to the constraint that it only chooses DCs that have a high probability of being safe (caution).  We also propose an extension, {\em SDF-Bayes-AR}, that accounts for patient heterogeneity and enables heterogeneous patient recruitment. Extensive experiments based on both synthetic and real-world datasets demonstrate the advantages of SDF-Bayes over state of the art DC trial designs in terms of accuracy and safety.
\end{abstract}

\section{INTRODUCTION}
\vspace{-0.05in}
The use of combinations of drugs is becoming an increasingly valuable--and common--treatment modality \citep{bamias2011report,flaherty2012combined,ocana2019efficacy,kelly2018oncology}, especially in the treatment of cancer, where it has been widely observed that combinations of drugs can be effective when single drugs are not \citep{paller2019factors}.  (Similar observations have been made for other diseases, including COVID-19.)  As a result, there has been an enormous effort to test and validate drug combinations for treatment; indeed combination trials now account for more than 25\% of all clinical trials in oncology \citep{wu2014characteristics}.

However, clinical trials of drug combinations face greater challenges than those of single drugs, especially in Phase I trials which are required to find safe doses.  The essential problem is that  the number of potential dose-combinations (DCs) to be tested in a trial increases \emph{exponentially} with the number of drugs, but the patient budget cannot scale proportionately.  {Indeed, a typical real-world Phase I trial often recruits fewer than 100 patients; a few examples are shown in Table \ref{table:trials}. The limited patient budget constrains the number of DCs that can be throughly tested. Moreover, because drugs interact differently with different body chemistry of different groups of patients  (e.g., with the different hormone balances and levels in males and females), it is frequently possible to identify in advance  groups of patients who might be expected to exhibit very different tolerances for the same DC  \citep{sun2015two,kim2009ugt1a1,dasari2013Phase,moss2015efficacy,wages2015Phase}.

In view of the growing importance of drug combinations in the treatment of disease, it is  of enormous importance to  design Phase I drug combination trials in a way that is efficient, informative and safe \citep{hamberg2010dose}. This design includes the path of DC testing and, in the presence of identified heterogeneous groups, the allocation of patient budgets to groups as well.  

\begin{table*}
\vspace{-0.1in}
	\caption{Examples of Phase I clinical trial studies with drug combinations}
	\label{table:trials}
	\footnotesize
	\fontsize{8pt}{8pt}\selectfont
	\begin{center}
		\setlength\tabcolsep{3pt}
		\begin{tabular}{ccccc}
			\toprule
			\textbf{Study} & \textbf{Drugs} & \textbf{No. of DC} & \textbf{No. of Patients} & \textbf{Target disease} \\
			\midrule
			\citep{plummer2008Phase} & Rucaparib \& temozolomide  & 12 & 32 & Advanced solid tumors \\
			\midrule
			\citep{bailey2009bayesian} & Nilotinib \& imatinib & 20 & 50 & \makecell{Gastrointestinal stromal tumors} \\
			\midrule
			\citep{bagatell2014Phase} & \makecell{Temsirolimus \& irinotecan \\\& temozolomide} & 24 & 71 & Solid tumors \\
			\midrule
			\citep{calvo2017Phase} & \makecell{Dacomitinib \& Figitumumab} & 12 & 74 & Advanced solid tumors \\
			\bottomrule
		\end{tabular}
	\end{center}
	\vspace{-0.15in}
\end{table*}

This paper develops a new dose-finding Phase I clinical trial method for drug combinations and heterogeneous patient groups and demonstrates that it is superior to existing methods.
For a given  patient budget, the design objective is to maximize the probability of finding the maximum tolerated dose (MTD), defined to be the DC that is closest to a given target toxicity threshold,  subject to constraints on the exposure of  patients  to unsafe doses throughout the trial.  Our safe dose-finding (SDF) method employs an adaptive design: the DC to be tested in the current round is chosen on the basis of past observations. This is possible because Phase I trials are not blind;  the trialist knows the DC given to each  patient and observes the outcome. The   \textit{SDF-Bayes} algorithm builds on a novel learning principle we call \textit{Cautious Optimism}, which manifests by combining two opposing ideas: (1) SDF-Bayes constrains the choice of DC to be tested in the current round to a set of  DCs that it estimates to be  unlikely to violate the safety constraint; this is the principle of {\em caution}, and (2) within this constrained set of DCs, SDF-Bayes chooses the DC to be tested in the current round to be the one estimated to be {\em most likely} to be the MTD; this is the \textit{optimistic belief principle}  \citep{aziz2019multi}.

To deal with settings in which potentially heterogeneous patient groups can be identified in advance, we also propose a extension, that we call SDF-Bayes-AR, in which both the DC to be tested {\em and} the patient group to be sampled in the current round are chosen adaptively.  To determine the group from which to recruit the next patient, SDF-Bayes-AR uses the criterion of {\em expected improvement} (EI).  Adaptive recruitment is especially useful when there is prior information about one or several groups  \citep{pallmann2018adaptive,park2018critical,atan2019sequential}, and an appealing feature of our approach is that it can smoothly incorporate prior information, be it from the drug development phase, from dose-toxicity models, from tests {\em in vitro} and in animals,  or from previous trials  \citep{gasparini2013general,shen2019harnessing}.\looseness=-1

We validate the proposed designs via extensive simulated trials using both synthetic and real-world datasets.   We  show that, using a realistic  number of patients, our algorithms provide significantly more accurate recommendations than state-of-the-art designs, while obeying the safety constraints,  for both homogeneous and heterogeneous patient populations.

\vspace{-8pt}

\section{DC-FINDING CLINICAL TRIALS}
\vspace{-0.1in}
\subsection{Dose-Toxicity Model}
\vspace{-0.05in}
We consider a dual-agent dose-finding Phase I clinical trial for the combination of agents (drugs) A and B. We assume discrete dose levels $\mathcal{J}\!=\!\{1,...,J\}$ for agent A and $\mathcal{K}\!=\!\{1,...,K\}$ for agent B.  We use $(j,k)$ to denote the combination of dose $j$ of agent A and dose $k$ of agent B so that the set of all DCs is just $\mathcal{A}\!=\!\mathcal{J}\times\mathcal{K}$.  We model the toxicity event $Y_{jk}$ of DC $(j,k)$ as a Bernoulli random variable with unknown parameter $p_{jk}$. We set $Y_{jk} = 1$ to indicate that a dose-limiting toxicity (DLT) was observed for $(j,k)$, and $Y_{jk} = 0$ otherwise.

We assume the toxicities  follow a parametric joint dose-toxicity model \vspace{-3pt}
\begin{equation}
\pi : \Theta \times \mathcal{J} \times \mathcal{K} \to [0,1]  \vspace{-3pt}
\end{equation}
where $\Theta$ is some space of parameters, and $\pi(\theta,j,k) = p_{jk}(\theta)$ is the  toxicity of the DC $(j,k)$ if the parameter is $\theta \in \Theta$.  The {\em true} vector of parameters $\theta^*$ is unknown and must be learned/estimated.  The literature has suggested various dose-toxicity models \citep{gasparini2013general,riviere2014bayesian}; we present details of some  commonly-used  models in the Supplementary Material; we focus here on developing a methodology that can be used with {\em many} parametric joint dose-toxicity models.

\subsection{Problem Formulation}
We consider an adaptive Phase I clinical trial for drug combinations with a given  patient budget $T$.  The nominal trial objective is to find the DC whose estimated toxicity is closest to the threshold $\xi$; i.e. to find any DC that belongs to
\vspace{-0.05in}
$$
 \mathcal{A}^* = \argmin_{jk}  \big| p_{jk}(\theta^*) - \xi \big|.  \vspace{-6pt}
$$
\vspace{-0.1in}

For each $t$, write
$\mathcal{O}(t)=\{(Y_\tau,a(\tau))\}^{t-1}_{\tau=1}$ for the history of  trial actions and observations {\em before} $t$; $a(\tau)$ is the DC that was administered to patient $\tau$ and $Y_\tau$ is the observed toxicity for  DC $a(\tau)$ on patient $\tau$. As in typical adaptive clinical trial designs, all DCs are actionable in the trial. A DC-finding algorithm $\Pi$ maps the history $\mathcal{O}(t)$ to the DC $a(t)\in \mathcal{A}$ that is to be administered to patient $t$.   At the end of the clinical trial, the algorithm recommends a DC  $\hat{a}^*$.  The recommendation is {\em correct} if
$\hat{a}^* \in \mathcal{A}^*$; otherwise the recommendation is an {\em error}. We set the objective of our algorithm to minimize the probability that errors occur. (Because the occurrence of toxicity events is stochastic, there is always some positive probability that errors will occur.)\looseness=-1

Because testing unsafe DCs should be avoided, we insist that, with high probability,  the average toxicity should not exceed the toxicity threshold plus a margin of error.  We formalize this by defining the   {\em  DLT  observation rate} to be\looseness=-1 \vspace{-6pt}
$$
S(T)=\frac{1}{T}\sum_{t=1}^T Y_t \vspace{-3pt}
$$
and we impose the {\em safety constraint} \vspace{-3pt}
\begin{equation}
	\label{eqn:safety_constraint}
	\mathbb{P}\big(S(T)\leq \xi+\epsilon\big) \geq 1-\delta, \vspace{-3pt}
\end{equation}
where $\epsilon>0$ is a prescribed  margin of error and $\delta > 0$ is a prescribed acceptable probability of failure.


\section{CAUTIOUS OPTIMISM}

\subsection{Preliminaries}
Given the dose-toxicity structure, we begin with a prior distribution $f(\theta)$ on the parameter vector $\theta$ and update the posterior distribution, based on the observations.  (The assumption of a prior distribution is common in the literature, and is natural, because the toxicities of the individual drugs are usually understood on the basis of prior use, and the various drug combinations will have already been tested in vitro and in animals.)  We write $f(\theta | \mathcal{O}(t))$ to denote the posterior distribution of $\theta$ following the history
$\mathcal{O}(t)$. The likelihood of $\theta$ in round $t$ for the observations $\mathcal{O}(t)$ satisfies
\[
L(\theta|\mathcal{O}(t))\propto \prod_{a\in\mathcal{A}}p_{a}(\theta)^{s_{a}(t)}(1-p_{a}(\theta))^{n_{a}(t)-s_{a}(t)},  \vspace{-3pt}
\]
where $n_a(t)$ is the number of times DC $a$ has been chosen before round $t$ and $s_a(t)=\sum_{\tau=1}^{t-1}\mathbb{I}[a(\tau)=a]Y_\tau$ is the number of  DLTs observed for the DC $a$ before round $t$.  The posterior distribution satisfies
\begin{equation}
	\label{eqn:posterior_dist}
	f(\theta | \mathcal{O}(t)) \propto L(\theta|\mathcal{O}(t)) f(\theta).  \vspace{-3pt}
\end{equation}

\subsection{Cautious Optimism in Bayesian SDF}
\label{sec:cautious_optimism}

We base our algorithm on the principle of ``optimism in the face of uncertainty''  \citep{Bubeck2012}, which means choosing the DC that is currently estimated to be most likely to be the MTD, but we maintain safety by constraining the set of DCs from which we choose.

\begin{paragraph}{Optimism for Efficiency}
	We assess which DC is most likely to be the MTD by using the posterior distribution $f(\theta|\mathcal{O})$ as described in \eqref{eqn:posterior_dist}.
	For this, in our algorithm we follow the literature (see \citep{riviere2014bayesian}, for example) and find the DC \vspace{-3pt}
	$$
	\argmax_{a} {\mathbb P}\big( \{ \theta : | p_{a}(\theta) - \xi | < u \} \big), \vspace{-3pt}
	$$
	where $u$ is some prescribed allowable margin of error, and the probability is taken with respect to the posterior distribution on the parameter space $\Theta$.

	  	The probability  $G_a^{\mathcal{O}(t)}(u)$ that the toxicity  of DC $a$ belongs to the given target toxicity interval $[\xi-u,\xi+u]$ is\looseness=-1
	\begin{IEEEeqnarray}{rl}
		\label{eqn:measure}
		G_a^{\mathcal{O}(t)}(u) ~&= \mathbb{P}[p_a(\theta)\in[\xi-u,\xi+u]|\mathcal{O}(t)]  \\
		& = \int_{\Theta}\mathbb{I}[p_a(\theta)\in[\xi-u,\xi+u]]f(\theta|\mathcal{O}(t))d\theta. \nonumber
	\end{IEEEeqnarray}
	Then, we find a DC  $a(t)$ that is deemed most likely to have toxicity in $[\xi-u,\xi+u]$; i.e.
	$$
	a_o(t) = \argmax_{a}G_a^{\mathcal{O}(t)}(u).
	$$
	The DC is chosen to be allocated if it is deemed ``safe enough'' (see the next paragraph) at round $t$.
	If the argmax is not a singleton, we choose arbitrarily, subject to maximizing the total dose of both drugs.
\end{paragraph}

\begin{paragraph}{Caution for Safety} To determine the set of DCs that are ``safe enough'' in round $t$ we first choose a hyperparameter $v > 0$ that controls conservativeness  and define
	\begin{align}
	F_a^{\mathcal{O}(\tau)}(v) &= \max\Big\{x\in[0,1]:\mathbb{P}\big[p_a(\theta)\leq x \, \big| \, \mathcal{O}(\tau)\big ]\leq v \Big\}
	           \nonumber \\
	\Phi(t,v) &= \sum_{\tau=1}^{t-1}F_{a(\tau)}^{\mathcal{O}(t)}(v) \nonumber\\
	r(t,v) & = (\xi+\epsilon)t - \Phi(t,v). \label{eqn:residual}
	\end{align}
 Roughly speaking, $\Phi(t,v)$ represents the number of DLT observations that would have been expected before round $t$, given the posterior. For safety, we are ``allowed'' a DLT observation rate $\xi + \epsilon$ so if
	$\Phi(t,v) \leq (t-1)(\xi + \epsilon)$  the safety constraint holds in expectation after $t-1$ rounds; if we choose $a(t)$ so that 	$\Phi(t+1,v) \leq t (\xi + \epsilon)$ then we will have met the  safety constraint in expectation after $t$ rounds.  We cannot be assured of choosing such an  $a(t)$  because we do not know what the posterior will be after $t$ rounds, but if the posterior after $t$ rounds were the same as the posterior after $t-1$ rounds and we choose $a(t)$ to have expected toxicity less than the {\em residual} $r(t,v)$  then the safety constraint will be met in expectation after $t$ rounds.  So define  \vspace{-3pt}
	\begin{equation}
		\label{eqn:residual_dc}
		\mathcal{A}_r(t) \coloneqq \{ a \in \mathcal{A}: F_a^{\mathcal{O}(t)}(v) \leq r(t,v)\}. \vspace{-3pt}
	\end{equation}
	Unfortunately it might be that $a_o(t)\notin\mathcal{A}_r(t)$. (That would certainly be the case if $r(t,v) < 0$, which means that the safety constraint had already been violated in expectation.)
	In that case we can use a {\em conservative} DC, one whose expected toxicity is no greater than $\xi$, so that after $t$ rounds the  residual $r(t+1,v)$ will be greater than the residual $r(t,v)$.
	The set of conservative DCs is  \vspace{-3pt}
	$$
	\mathcal{A}_{c}(t) \coloneqq \lbrace a \in\mathcal{A} :
	F_a^{\mathcal{O}(t)}(v)\leq \xi \rbrace.    \vspace{-3pt}
	$$
	We allocate the most likely conservative DC to be the MTD; i.e., $\argmax_{a\in\mathcal{A}_c(t)}G_a^{\mathcal{O}(t)}(u)$.
	Finally, if $\mathcal{A}_{c}(t) = \emptyset$ we terminate the procedure with no recommendation.

		 \end{paragraph}

	\begin{paragraph}{Theoretical Guarantee}
	The following Proposition shows that, when this cautiously optimistic procedure can be carried out, it maintains the safety constraint by keeping the residual non-negative with a high probability.  (The proof is in the Supplementary Materials.)

	\vspace{-3pt}
	\begin{proposition}
		\label{prop:1}
		Fix $t \leq T$.  For each $\tau \leq t$, set  $v_\tau=(1-\delta)^{1/\tau}$.  If the residuals $r(\tau, v_\tau)$ are non-negative for all $\tau \leq t$ then the cautious optimism principle in Bayesian SDF satisfies
		$$
		\mathbb{P}\left[\left.\frac{1}{t}\sum_{\tau=1}^t p_{a(\tau)}(\theta)\leq \xi+\epsilon \right| \mathcal{O}(t)\right]\geq 1-\delta.
		$$
		\vspace{-9pt}
	\end{proposition}
\end{paragraph}

\subsection{Algorithm Description}
Here we provide an implementation of the  \emph{cautious optimism} principle described above.  Because we are considering  an arbitrary joint dose-toxicity model,  our implementation relies on  Bayesian sampling in order to ensure the universal applicability of the algorithm. We comment that our algorithm can be easily extended for drug combinations with more than two drugs if a corresponding dose-toxicity model is given.

We denote the number of samples from the posterior distribution $f(\theta|\mathcal{O})$ by $L$, and the samples in round $t$ as $\tilde{\Theta}(t)=\{\theta^{(l)}(t)\}_{l\in[L]}$.
We use a Gibbs sampler \citep{gilks1995adaptive}; this is a common multidimensional Bayesian sampling algorithm.  Inside the Gibbs sampler, we use the adaptive rejection Metropolis sampling method \citep{gilks1995adaptive}. Details can be found in the Supplementary Material.

In round $t$, we  use the samples to approximate the probability $G_a^{\mathcal{O}(t)}(u)$ by  \vspace{-5pt}
\[
\tilde{G}_a^{\mathcal{O}(t)}(u)=\frac{1}{L}\sum_{l=1}^L \mathbb{I} \big\{ t : \xi-u \leq p_a(\theta^{(l)}(t)) \leq \xi+u \big \}.  \vspace{-5pt}
\]
Using this approximation, the DC most likely to be the MTD in round $t$ is $\tilde{a}_o(t)=\argmax_{a\in\mathcal{A}} \tilde{G}_a^{\mathcal{O}(t)}(u)$.
SDF-Bayes then infers whether the safety constraint is violated or not for the chosen DC by evaluating the residual.
To calculate the residual in practice, we define $\tilde{F}_a^{\mathcal{O}(t)}(v)\coloneqq\texttt{Prctile}(a,\tilde{\Theta}(t),v)$ that returns the percentile of the toxicities of DC $a$ calculated from the samples $\tilde{\Theta}(t)$ for the percentage $v\in[0,1]$.
Then, from \eqref{eqn:residual}, we calculate the residual $r(t,v)$ in round $t$ by
$
r(t,v)=(\xi+\epsilon) t - \sum_{\tau=1}^{t-1} \tilde{F}_{a(\tau)}^{\mathcal{O}(t)}(v)
$
and define $\tilde{\mathcal{A}}_r(t)$ as in \eqref{eqn:residual_dc} with $\tilde{F}_a^{\mathcal{O}(t)}(v)$.
To keep the residual non-negative, SDF-Bayes accepts the chosen DC $\bar{a}(t)$
if it does not make the residual negative:
$\tilde{a}_o(t)\in\tilde{\mathcal{A}}_r(t)$.
Otherwise, it rejects the chosen DC and chooses the most likely DC to be the MTD in the set of conservative DCs; i.e., $\argmax_{a\in\tilde{\mathcal{A}}_c(t)} \tilde{G}_a^{\mathcal{O}(t)}(u)$, where
\[
\tilde{\mathcal{A}}_c(t,v)= \lbrace a\in\mathcal{A}:\tilde{F}_a^{\mathcal{O}(t)}(v)\leq \xi \rbrace.
\]
If the set is empty, then the trial is terminated in the cautious optimism principle. However, this may be too conservative in the practical implementation, especially if $v$ is relatively large, because it implies that there is no ``safe enough'' DC in a conservative view, not all DCs are unacceptably unsafe. Hence, if $\tilde{\mathcal{A}}_c(t,v)=\emptyset$, SDF-Bayes first finds $w$ so that $\tilde{\mathcal{A}}_c(t,w)\neq\emptyset$ and $\tilde{\mathcal{A}}_c(t,v')=\emptyset$ for all $v'>w$. Then, it continues the trial by choosing the most likely DC in $\tilde{\mathcal{A}}_c(t,w)$ if $w>\psi$, where $\psi$ is a predefined parameter for early termination rule. Otherwise, it terminates the trial with no recommendation because it implies all DCs are unsafe.
(This early termination rule is widely used in the literature \citep{riviere2014bayesian,yin2009bayesian}.)
%
With the chosen DC $a(t)$, the DLT $Y_t$ is then observed, and the observation $\mathcal{O}(t+1)$
is updated. This process repeats until $T$ patients have been administered.
At the end of the clinical trial, the DC recommendation is given by
$\hat{a}^*=\argmax_{a\in\mathcal{A}}\tilde{G}_a^{\mathcal{O}(T)}(u)$.
The pseudocode of the algorithm is summarized in Algorithm 1.

\setlength{\textfloatsep}{6pt}
\begin{algorithm}[t]
	\caption{\textsc{SDF-Bayes}}
	\label{alg:safe_dc_finding_bayes}
	\begin{algorithmic}[1]
		\While{$t\leq T$}
		\State Sample $\theta$ from their posterior distribution
		\If{$\tilde{a}_o(t) \in\tilde{\mathcal{A}}_r(t)$}
		\State $a(t)\leftarrow \tilde{a}_o(t)$
		\ElsIf{$\tilde{\mathcal{A}}_c(t,v)\neq\emptyset$}
		\State $a(t)\leftarrow \argmax_{a\in\tilde{\mathcal{A}}_c(t,v)} \tilde{G}_a^{\mathcal{O}(t)}(u)$
		\Else
		\If{$w>\psi$}
		\State $a(t)\leftarrow \argmax_{a\in\tilde{\mathcal{A}}_c(t,w)} \tilde{G}_a^{\mathcal{O}(t)}(u)$
		\Else
		\State Terminate trial without recommendation
		\EndIf
		\EndIf
		\State Observe the DLT $Y_t$
		\State Update $s_{a(t)}(t+1)$ and $n_{a(t)}(t+1)$
		\State $t\leftarrow t+1$
		\EndWhile
		\State \textbf{Output:} $\hat{a}^*=\argmax_{a\in\mathcal{A}}\tilde{G}_a^{\mathcal{O}(T)}(u)$
	\end{algorithmic}
\end{algorithm}

\section{HETEROGENEOUS GROUPS}
\subsection{Problem formulation}
We now show how to adapt SDF-Bayes to deal with heterogeneous groups.  We continue to assume a total budget of $T$ patients, which now can be distributed across $M$ patient groups  $\mathcal{M}=\{1,2,...,M\}$.  We allow for the possibility that the dose-toxicity model varies across different groups, and model the toxicity $Y_{jk}^m$ of $(j,k)$ for group $m$ as a Bernoulli random variable with unknown parameter $p_{jk}^m$.  For each group $m$, we assume $p_{jk}^m$ follows a parametric joint dose-toxicity model  $p_{jk}^m(\theta)=\pi^m(j,k,\theta)$; we write
$\theta_m^*$ for the true parameter and $\xi_m$ for the prescribed toxicity threshold. For patient $t$, the algorithm first chooses a group $g(t) \in  \mathcal{M}$ from which to recruit the next patient, then chooses a DC for the recruited patient based on the history $\mathcal{O}(t)=\{(Y_\tau,g(\tau),a(\tau)\}_{\tau=1}^{t-1}$ prior to round $t$.  The outcome (DLT or not) is then observed and recorded. This continues until the total budget $T$ is exhausted.

At the end of the  trial, the algorithm recommends, for each group $m$, a  DC $\hat{a}^*_m$ to be used as the MTD for that group.
As before, the (set of) true MTD(s) for group $m$ is \vspace{-3pt}
$$
\mathcal{A}_m^*=\argmin_{a\in\mathcal{A}}|p_a^m(\theta_m^*)-\xi_m|, \vspace{-3pt}
$$
and the recommendation is an {\em error} if $\hat{a}^*_m\notin\mathcal{A}^*_m$.
The safety constraint for  group $m$ is
$\mathbb{P}[S_m(T)\leq\xi_m+\epsilon]\leq 1-\delta,~\forall m\in\mathcal{M}$, where
$S_m(T)=\frac{\sum_{t=1}^T Y_t \mathbb{I}[g(t)=m]}{\sum_{t=1}^T \mathbb{I}[g(t)=m]}$.

\subsection{SDF-Bayes for Heterogeneous Groups}
The goal of a clinical trial with heterogeneous groups is to minimize the DC recommendation error while satisfying the safety constraints of {\em each} group.  Given the flexibility of recruiting patients from different groups, it is intuitive that {\em uniform} recruitment across groups, which does not utilize the history information prior to each decision time, may be inefficient.  For example, if a particular patient group has already accumulated sufficient observations to determine the MTD with high confidence, recruiting more patients for this group is not as beneficial as for other groups.
We reinforce this intuition by proposing SDF-Bayes-AR, a modification of SDF-Bayes in which patients are \textit{adaptively} recruited.  We will prove that SDF-Bayes-AR utilizes the limited number of patients more efficiently, by adaptively recruiting patients to obtain the best information for each group while satisfying the safety constraints.

\begin{paragraph}{Adaptive Patient Recruitment}
	In SDF-Bayes-AR, we use the probability $G_a^{\mathcal{O}(t)}(u)$ in \eqref{eqn:measure} to measure the likelihood of MTD.
	We denote the probability in round $t$ that the DC $a$ for group $m$  is the MTD by
	$$
	G_{m,a}^{\mathcal{O}(t)}(u) = \int_{\Theta}\mathbb{I}[p_{a}^m(\theta)\in[\xi_m-u,\xi_m+u]]f_m(\theta|\mathcal{O}(t))d\theta,
	$$
	where $f_m(\theta|\mathcal{O}(t))$ is the posterior distribution of $\theta$ for group $m$.
	The posterior distribution for each group can be calculated as in \eqref{eqn:posterior_dist} by using, for each DC $a$,  the number ($n^m_{a}(t)$) of times $a$ is used for group $m$ before round $t$ and the number ($s^m_{a}(t)$) of DLTs  observed when  $a$ is  used within group $m$ before round $t$.  The improvement of the probability  of the most probable DC for group $m$ in round $t$ for an \textit{additional} patient with DC $a'$ and observed toxicity $Y$ is given by
	$$
	I_{m,a'}^{\mathcal{O},Y}(u)=\left|G_{m,*}^{\{\mathcal{O},(Y,m,a')\}}(u)-G_{m,*}^{\mathcal{O}}(u)\right|,
	$$
	where $G_{m,*}^{\mathcal{O}}=\max_{a\in\mathcal{A}}G_{m,a}^{\mathcal{O}}$.
	Define the \textbf{expected improvement (EI)}
	\begin{align*}
	H_{m,a'}^{\mathcal{O}(t)}(u) &= \int_{\Theta} \Big\{ p_{a'}^m(\theta)I_{m,a'}^{\mathcal{O}(t),1}(u)  \\
	& \hspace{30pt} + (1-p_{a'}^m(\theta)I_{m,a'}^{\mathcal{O}(t),0}(u) \Big\} f_m(\theta|\mathcal{O}(t))d\theta
	\end{align*}
	Denote the tentatively allocated DC of group $m$ in round $t$ by $a_m(t)$.
	Then, in SDF-Bayes-AR, we adaptively recruit a patient of group $m^*$ by using $a_m(t)$'s as
	$
	m^*(t) = \argmax_{m\in\mathcal{M}}H^{\mathcal{O}(t)}_{m,a_m(t)}(u).
	$
\end{paragraph}

\begin{paragraph}{Algorithm Description}
	In round $t$, we apply SDF-Bayes for each group $m$ to find the  DC $a_m(t)$ that should be chosen for that group.  Specifically, we use lines 2--13 of Algorithm \ref{alg:safe_dc_finding_bayes}.
	We then approximate the EI $H_{m,a_m(t)}^{\mathcal{O}(t)}(u)$ for group $m$.
	To this end, we  sample from three different posterior distributions, $f_m(\theta|\mathcal{O}(t))$, $f_m(\theta|\{\mathcal{O}(t),(1,m,a')\})$, and $f_m(\theta|\{\mathcal{O}(t),(0,m,a')\})$.
	We denote the sets of the samples by $\tilde{\Theta}_m(t)$, $\tilde{\Theta}_m^{(1,m,a')}(t)$, and $\tilde{\Theta}_m^{(0,m,a')}(t)$, respectively.
	We approximate the EI $H_{m,a'}^{\mathcal{O}(t)}(u)$ by
	$
	\tilde{H}_{m,a'}^{\mathcal{O}(t)}(u)=\tilde{p}_{a'}^m(\mathcal{O}(t)) \tilde{I}^{\mathcal{O}(t),1}_{m,a'}(u)
	+ (1-\tilde{p}_{a'}^m(\mathcal{O}(t)))\tilde{I}^{\mathcal{O}(t),0}_{m,a'}(u),
	$
	where $\tilde{p}_a^m(\mathcal{O}(t))=\frac{1}{L}\sum_{\theta\in\tilde{\Theta}_m(t)} p_a^m(\theta)$.
	Finally, the group to be recruited is chosen to be
	$g(t)=\argmax_{m\in\mathcal{M}}\tilde{H}_{m,a_m(t)}^{\mathcal{O}(t)}(u)$, and the DC to be allocated to that group is $a(t)=a_{g(t)}(t)$.  (Ties for $g(t)$ or $a_{g(t)}$ are broken arbitrarily.)  We then observe  the DLT $Y_t$ from the recruited patient of group $g(t)$ with the allocated DC $a(t)$ and construct the history $\mathcal{O}(t+1)$.
	At the end of the trial, the DC recommendation for each group is given by
	$\hat{a}^*_m=\argmax_{a\in\mathcal{A}}\tilde{G}_{m,a}^{\mathcal{O}(T)}(u)$.
	A more detailed description of SDF-Bayes-AR is provided in the Supplementary Material.  \looseness=-1
\end{paragraph}

\section{EXPERIMENTS}
\label{sec:experiments}
Here we describe a variety of experiments using a real-world  dataset and several synthetic data sets.  Synthetic datasets are widely used in designing and evaluating novel Phase I clinical trials in order to thoroughly investigate the design before subjecting human subjects to a potentially dangerous regime of drugs.  Moreover, because a Phase I trial never identifies toxicity probabilities exactly and never identifies the MTD with certainty, a real-world dataset is not  ``completely realistic'' either.\looseness=-1

\begin{table*}[t]
	\fontsize{8pt}{8pt}\selectfont
	\centering
	\caption{True toxicity probabilities in datasets A,B,C,D,RW}
	\label{table:trialdatasets}
	\vspace{0.1in}
	\setlength\tabcolsep{1.5pt}
	\begin{tabular}{c|cccc|cccc|cccc|cccc|cccc}
		\toprule
		&           \multicolumn{4}{c|}{Synthetic A}           &  \multicolumn{4}{c|}{Synthetic B}  &  \multicolumn{4}{c|}{Synthetic C}  &           \multicolumn{4}{c|}{Synthetic D}           &       \multicolumn{4}{c}{Real-World}        \\ \midrule
		~~~~~ &  1   &       2       &       3       &       4       &  1   &  2   &  3   &       4       &  1   &  2   &       3       &  4   &       1       &  2   &       3       &       4       &  1   &  2   &       3       &       4       \\ \midrule
		3   & 0.15 & \textbf{0.30} &     0.45      &     0.50      & 0.09 & 0.12 & 0.15 & \textbf{0.30} & 0.08 & 0.15 &     0.45      & 0.60 & \textbf{0.30} & 0.42 &     0.52      &     0.62      & 0.13 & 0.21 & \textbf{0.30} &     0.43      \\
		2   & 0.10 &     0.15      & \textbf{0.30} &     0.45      & 0.05 & 0.10 & 0.13 &     0.15      & 0.05 & 0.12 & \textbf{0.30} & 0.55 &     0.10      & 0.20 & \textbf{0.30} &     0.40      & 0.08 & 0.13 &     0.20      & \textbf{0.30} \\
		1   & 0.05 &     0.10      &     0.15      & \textbf{0.30} & 0.02 & 0.08 & 0.10 &     0.11      & 0.02 & 0.10 &     0.15      & 0.50 &     0.05      & 0.12 &     0.20      & \textbf{0.30} & 0.04 & 0.07 &     0.11      &     0.17      \\ \bottomrule
	\end{tabular}
\end{table*}

The real-world dataset RW we use is taken from \citep{bailey2009bayesian}, which reports the observations  during a real Phase I clinical trial for the combination of two oncology drugs, nilotinib and imatinib. \citet{bailey2009bayesian} constructs a dose-toxicity model based on  DLT observations and on prior information about the drugs.  The dose combinations consist of 400, 600, and 800 mg of nilotinib (drug A) and 0, 400, 600, and 800 mg of imatinib (drug B). (Further details are in the Supplementary Materials.)  The synthetic datasets A,B,C,D are constructed as variations on RW.  (Additional synthetic datasets and results are presented in the Supplementary Materials.)  Table \ref{table:trialdatasets} reports the true toxicities for the various doses in the datasets A,B,C,D,RW; the true MTD's are shown in boldface.

For comparison purposes, we implemented five dose-finding algorithms: SDF-Bayes; DF-Bayes (SDF-Bayes without caution, chosen to illustrate the effect of optimism without caution); SOTA Bayes  \citep{riviere2014bayesian}, a state-of-the-art Bayesian dose-finding algorithm;
IndepTS \citep{aziz2019multi}, a Thompson sampling-based multi-armed bandit (MAB) clinical trial algorithm; and StructMAB, a structured MAB-based clinical trial algorithm that exploits the joint dose-toxicity model based on the structured bandit  method from  \citep{gupta2019unified} while taking into account the safety constraint, which is an advanced version of the safe dose allocation method in \cite{shen2020icml} to drug combinations.
In the algorithms that exploit the joint dose-toxicity model (SDF-Bayes, DF-Bayes, SOTA Bayes and StructMAB), we use the logistic model in \cite{riviere2014bayesian} for fair comparison.\looseness=-1

\setlength{\intextsep}{6pt}%
\begin{table*}[h!]
	\fontsize{7pt}{7pt}\selectfont
	\caption{Safety Violations and Recommendation Errors  for datasets A,B,C,D,RW}
	\label{table:performance}
	\centering
	\vspace{0.1in}
	\setlength\tabcolsep{0.3pt}
	\begin{tabular}{c||c|c||c|c||c|c||c|c||c|c}
		\toprule
		&       \multicolumn{2}{c||}{Synthetic A}        &    \multicolumn{2}{c||}{Synthetic B}     &       \multicolumn{2}{c||}{Synthetic C}        &       \multicolumn{2}{c||}{Synthetic D}        &         \multicolumn{2}{c}{Real-World}         \\ \midrule
		Algorithms &     Safety Viol.      &    \makecell{Rec. Errors}    &  Safety Viol.   &    \makecell{Rec. Errors}    &     Safety Viol.      &    \makecell{RecErrors}    &     Safety Viol.      &    \makecell{Rec Errors}    &     Safety Viol.      &    \makecell{Rec Errors}    \\ \midrule
		SDF-Bayes  &    0.019$\pm$.004    & \textbf{0.205$\pm$.011} & 0.001$\pm$.001 &     0.193$\pm$.009      &    0.010$\pm$.002    & \textbf{0.443$\pm$.012} &    0.040$\pm$.005    & \textbf{0.344$\pm$.011} &    0.003$\pm$.001    & \textbf{0.368$\pm$.011} \\
		DF-Bayes  & \vio{0.411$\pm$.014} &     0.237$\pm$.012      & 0.020$\pm$.003 & \textbf{0.123$\pm$.008} & \vio{0.434$\pm$.012} &     0.643$\pm$.011      & \vio{0.460$\pm$.012} &     0.317$\pm$.011      & \vio{0.154$\pm$.008} &     0.341$\pm$.011      \\
		SOTA Bayes &    0.023$\pm$.004    &     0.233$\pm$.012      & 0.000$\pm$.000 &     0.196$\pm$.009      &    0.013$\pm$.003    &     0.570$\pm$.012      &    0.041$\pm$.005    &     0.394$\pm$.011      &    0.002$\pm$.001    &     0.391$\pm$.011      \\
		StructMAB  &    0.018$\pm$.004    &     0.516$\pm$.014      & 0.000$\pm$.000 &     0.488$\pm$.012      &    0.011$\pm$.002    &     0.740$\pm$.010      &    0.060$\pm$.006    &     0.564$\pm$.012      &    0.002$\pm$.001    &     0.621$\pm$.011      \\
		IndepTS   &    0.013$\pm$.003    &     0.603$\pm$.014      & 0.000$\pm$.000 &     0.847$\pm$.008      &    0.006$\pm$.002    &     0.819$\pm$.009      &    0.072$\pm$.006    &     0.623$\pm$.011      &    0.000$\pm$.000    &     0.749$\pm$.010      \\ \bottomrule
	\end{tabular}
\end{table*}

\vspace{-0.5em}
\subsection{Homogeneous Groups}
\vspace{-0.5em}

For each of these datasets, we conducted 5,000 runs of each algorithm, each with a pool of 60 patients, using
$\xi = 0.30, \epsilon = 0.05, \delta = 0.05$ (these values are typical of actual Phase I trials).
In any run, we counted a recommendation error if the recommended DC is not the true MTD and a safety violation if the DLT observation rate exceeds the threshold $\xi+\epsilon$.
 Table \ref{table:performance} reports, for each algorithm and each dataset, the proportions of  runs in which there was a safety violation and  runs in which the recommended DC was an error, with 95\% confidence intervals. Safety violations do not satisfy the safety constraint (i.e., the proportion of runs in which there was a safety violation is more than 0.05) are shown in red; in the absence of those safety violations, the best error performance for each dataset is shown in boldface.  As can be seen, for {\em every} dataset, SDF-Bayes satisfies the safety constraint while  making fewer recommendation errors than SOTA Bayes, Indep TS, or StructMAB.  Indeed, SOTA Bayes is the only one of these algorithms that is at all competitive with SDF-Bayes; Indep TS and StructMAB are recommending the wrong MTD more than half the time.  DF-Bayes makes fewer recommendation errors than SDF-Bayes for  dataset B, in which {\em no DC exceeds the threshold $\xi$},  but makes wildly unacceptable proportions of safety violations for all other datasets.

SDF-Bayes does best because cautious optimism allows it to more efficiently explore the boundary between safe and unsafe DCs.  In the Supplementary Materials, we document that SDF-Bayes is more often testing DC's that are believed to be close to being unsafe, while other algorithms more often test DC's that are believed to be safe.  Because the primary objective of a Phase I trial is to find the MTD while maintaining acceptable patient safety, which is determined by the trialist, SDF-Bayes is making the proper trade-off between accurate prediction and safety of patients in the trial.

\subsection{Heterogeneous Groups}
To evaluate SDF-Bayes for heterogeneous groups, we use the groups A, B whose toxicities are given by Synthetic A and Synthetic B, respectively, provided in Table \ref{table:trialdatasets}.  Because we have two groups, we allow for a total of $80$ patients.  For each simulated trial, we compute both the safety violations for each group and the overall safety violation for the whole trial.  (Other details are in the Supplementary Material.)
\looseness=-1

To evaluate the effectiveness of adaptive recruitment in SDF-Bayes-AR, we use as baselines the various Bayesian algorithms with uniform recruitment (UR), so that patients from each group are recruited with equal probability, and SOTA Bayes-AR in which the proposed adaptive patient recruitment is adopted to SOTA Bayes.  (We have already found that the MAB-based algorithms are not competitive, so we do not use them here.)  We also  apply the Bayesian algorithms to the entire population, treated as a single group called  EP, whose  true toxicity probabilities are just the averages of the toxicity probabilities for the two groups A and B.
At the end of each trial, the algorithms recommend a {\em single} DC for EP, which we evaluate as a recommendation for each group separately.  As before, we use boldface to indicate the best performance subject to satisfying the safety constraints.\looseness=-1

\begin{table*}[h!]
	\vspace{-0.1in}
	\caption{Safety Violations and Recommendation Errors}
	\label{table:performances_het_to_homo}
	\fontsize{8pt}{8pt}\selectfont
	\begin{center}
		\setlength\tabcolsep{2pt}
		\def\arraystretch{1.2}
		\begin{tabular}{c||c|c|c|c|c|c}
			\toprule
			\multirow{2}{*}{Algorithm} &          \multicolumn{3}{c|}{Safety Violation Rate}          &                \multicolumn{3}{c}{Recommendation Error Rate}                \\
			\cmidrule{2-7}        &     Entire trial     &     ~~Group A~~      &  ~~Group B~~   &        ~~Average~~        &         Group A         &         Group B         \\ \midrule
			SDF-Bayes-AR         &    0.000$\pm$.000    &    0.049$\pm$.006    & 0.002$\pm$.001 & \textbf{0.283$\pm$.012} & \textbf{0.287$\pm$.013} & \textbf{0.279$\pm$.012} \\
			SDF-Bayes-UR         &    0.000$\pm$.000    &    0.028$\pm$.005    & 0.001$\pm$.001 &     0.288$\pm$.013      & \textbf{0.287$\pm$.013} &     0.288$\pm$.013      \\
			DF-Bayes-UR         & \vio{0.095$\pm$.008} & \vio{0.452$\pm$.014} & 0.025$\pm$.004 &     0.257$\pm$.012      &     0.325$\pm$.013      &     0.188$\pm$.011      \\
			SOTA Bayes-AR       & 0.000$\pm$.000     & \vio{0.064$\pm$.006} & 0.000$\pm$.000 &     0.294$\pm$.011      &     0.301$\pm$.011      &    0.287$\pm$.011       \\
			SOTA Bayes-UR        &    0.000$\pm$.000    &    0.036$\pm$.005    & 0.000$\pm$.000 &     0.298$\pm$.013      &     0.314$\pm$.013      &     0.283$\pm$.012      \\ \midrule
			SDF-Bayes-EP         &    0.004$\pm$.002    & \vio{0.734$\pm$.012} & 0.000$\pm$.000 &     0.806$\pm$.011      &     0.706$\pm$.008      &     0.906$\pm$.008      \\
			DF-Bayes-EP         & \vio{0.139$\pm$.010} & \vio{0.842$\pm$.010} & 0.007$\pm$.002 &     0.819$\pm$.011      &     0.807$\pm$.011      &     0.830$\pm$.010      \\
			SOTA Bayes-EP        &    0.003$\pm$.002    & \vio{0.493$\pm$.014} & 0.000$\pm$.000 &     0.800$\pm$.011      &     0.692$\pm$.013      &     0.908$\pm$.008      \\ \bottomrule
		\end{tabular}
	\end{center}
	\vskip -0.15in
\end{table*}

\vspace{-0.1in}
\begin{paragraph}{Safety Violations and Recommendation Errors}
	Table \ref{table:performances_het_to_homo} shows  that SDF-Bayes-AR achieves the lowest overall error rate among the algorithms that satisfy the safety constraint, but that the improvement of  SDF-Bayes-AR over  SDF-Bayes-UR is marginal.  As we will see below, this is because no prior information is used; as we show, adaptive recruitment is more effective when prior information is available.

	When we force the algorithms to treat EP as a homogeneous group, we see that the rate of safety violations in group B is extremely low; as a result, the overall rate of safety violations is also low (for SDF-Bayes and SOTA Bayes, applied to EP) although the rate of safety violations for group A is extremely high.  We also see that, in this setting,  the recommendation errors are all very high.  This is because according to the averages of the toxicity probabilities for the two groups A and B, the true MTDs for EP are  (2,4) and (3,3), neither of which is an MTD for either group A or group B.  This highlights the danger of treating heterogeneous populations as if they were homogeneous.\looseness=-1
\end{paragraph}

\begin{table*}[h!]
	\caption{Impact of Prior Information (A: Group A, B: Group B, E: Entire trial)}
	\label{table:varying_prior}
	\fontsize{8pt}{8pt}\selectfont
	\begin{center}
		\setlength\tabcolsep{1.5pt}
		\def\arraystretch{1.2}
		\begin{tabular}{c|c||c|c|c|c|c|c|c|c|c}
			\toprule
			\multicolumn{2}{c||}{} & \multicolumn{3}{c|}{Fraction Recruited from Group} & \multicolumn{3}{c|}{Safety Violation Rates} & \multicolumn{3}{c}{Recommendation Errors} \\
			\cmidrule{3-11}
			\multicolumn{2}{c||}{} & $T_p=20$ & $T_p=40$ & $T_p=60$ & $T_p=20$ & $T_p=40$ & $T_p=60$ & $T_p=20$ & $T_p=40$ & $T_p=60$ \\
			\midrule
			\multirow{3}{*}{AR}
			& A & 0.517$\pm$.001 & 0.599$\pm$.003 & 0.719$\pm$.004 & 0.027$\pm$.004 & 0.014$\pm$.003 & 0.014$\pm$.003 & 0.286$\pm$.013 & 0.281$\pm$.012 & 0.258$\pm$.012  \\
			& B & 0.483$\pm$.001 & 0.401$\pm$.003 & 0.281$\pm$.004 & 0.047$\pm$.006 & 0.105$\pm$.009 & 0.207$\pm$.011 & 0.233$\pm$.012 & 0.170$\pm$.010 & 0.131$\pm$.009  \\
			& E  & - & - & - & 0.001$\pm$.001 & 0.002$\pm$.001 & 0.001$\pm$.001 & 0.259$\pm$.012 & 0.226$\pm$.012 & 0.195$\pm$.011  \\
			\midrule
			\multirow{3}{*}{UR}
			& A & 0.500$\pm$.000 & 0.500$\pm$.000 & 0.500$\pm$.000 & 0.031$\pm$.005 & 0.031$\pm$.005 & 0.031$\pm$.005 & 0.303$\pm$.013 & 0.303$\pm$.013 & 0.303$\pm$.013 \\
			& B & 0.500$\pm$.000 & 0.500$\pm$.000 & 0.500$\pm$.000 & 0.021$\pm$.004 & 0.039$\pm$.005 & 0.055$\pm$.006 & 0.234$\pm$.012 & 0.162$\pm$.010 & 0.127$\pm$.009 \\
			& E  & - & - & - & 0.006$\pm$.002 & 0.011$\pm$.003 & 0.017$\pm$.004 & 0.268$\pm$.012 & 0.232$\pm$.012 & 0.215$\pm$.011  \\

			\bottomrule
		\end{tabular}
	\end{center}
	\vskip -0.15in
\end{table*}

\begin{paragraph}{Impact of Prior Information}
	As we have noted, useful prior information about one or both groups may be available  \citep{hobbs2011hierarchical}.  To illustrate the impact of  prior information on the adaptive recruitment in SDF-Bayes-AR and hence on the results, we assume that the prior information comes from a
	previous trial  for group B; we parametrize the amount/quality of prior information by controlling the number of patients $T_p$ in the previous trial.  	Table \ref{table:varying_prior}  records the fraction of patients recruiting in group A, the safety violation rates, and the recommendation error rates
	of SDF-Bayes-AR and SDF-Bayes-UR for various sizes of prior trials (and hence various amounts of prior information).  We can see that SDF-Bayes-AR adaptively recruits the patients according to the amount of the prior information.
	In SDF-Bayes-AR,  patients from group B are seldom recruited after the most likely MTD DC for group B is determined with high probability.  (See the  Supplementary Material for more detail.)  Because patients from group A are recruited more frequently, more is learned about group A and the error rate for group substantially reduced; the cost is only a marginal increase in the error rate for group B.  As a result, the overall error rate is improved as expected.  Because adaptive recruitment  reduces the number of patients recruited from group B, it also reduces the possibility of balancing the risk of group B and increases the {\em rate} of safety violations in group B -- although {\em fewer} group B patients are exposed to DCs that are found to be unsafe. Overall, SDF-Bayes-AR outperforms SDF-Bayes-UR in both accuracy and safety: it makes fewer total errors with fewer total safety violations.  \looseness=-1

\end{paragraph}

\begin{table*}[h!]
	\caption{Comparison of adaptive clinical trial Phase I methodologies}
	\label{table:relatedwork}
	\footnotesize
	\fontsize{8pt}{8pt}\selectfont
	\begin{center}
		\setlength\tabcolsep{3pt}
		\begin{tabular}{c|c|c|c|c|c}
			\toprule
			\textbf{Reference} & \textbf{Principle} & \makecell{ \textbf{No. of DCs} } &  \makecell{\textbf{Toxicity}\\ \textbf{Model}} & \textbf{Safety}  & \makecell{\textbf{Heterogeneous}\\\textbf{Groups}} \\
			\hline
			\makecell{ \cite{villar2015multi,garivier2017thresholding};\\ \cite{villar2018covariate} } & MAB  &  less than 10 & No & No & No \\ \hline
			\cite{varatharajah2018contextual} & MAB & less than 10 & No & No & Yes \\ \hline
			\cite{aboutalebi2019learning} & MAB &  less than 10 & No & Implicitly considered & No \\ \hline
			\cite{aziz2019multi} & MAB &  less than 10 & Fixed & Implicitly considered & No \\ \hline
			\cite{shen2020icml} & MAB &  less than 10 & Fixed & $\delta$-safety guaranteed & No \\ \hline
			\cite{lee2020contextual} & MAB & less than 10 & Fixed & $\delta$-safety guaranteed & Yes \\ \hline
			\cite{yan2017keyboard} & Bayesian &  less than 10& No &  Dose-escalation  & No \\ \hline
			\cite{wages2015Phase} & Bayesian & less than 10& Fixed & Dose-escalation  & Yes \\ \hline
			\cite{yin2009bayesian,riviere2014bayesian} & Bayesian &  more than 10 & Fixed &  Dose-escalation  & No \\ \hline\hline
			This work (SDF-Bayes) & Bayesian &  more than 10 & Arbitrary & $\delta$-safety guaranteed & Yes \\
			\bottomrule
		\end{tabular}
	\end{center}
	\vspace{-0.1in}
\end{table*}

\section{RELATED WORK}
\begin{paragraph}{Bayesian methodology for clinical trials}
Bayesian methodology has been widely used for clinical trials, first to  label the effectiveness of treatments
\citep{atan2019sequential,berry2006bayesian}.  It has also been used  in  dose-finding clinical trials \citep{wages2015Phase,o1990continual} and for drug combination trials \citep{riviere2014bayesian,yin2009bayesian,yin2009latent,yan2017keyboard}, which are about online learning for dose-finding clinical trials.  In the latter setting it is typically  used in conjunction with a dose-toxicity model because it can significantly reduce the size of the search space \citep{shen2019harnessing}.
By using the posterior distribution of the parameters of the dose-toxicity model, traditional Bayesian DC-finding algorithms are proposed to find the MTD based on dose escalation and de-escalation from the lowest dose levels for safety \citep{riviere2014bayesian,yin2009bayesian,yan2017keyboard}.\looseness=-1
\end{paragraph}

\begin{paragraph}{MAB for clinical trials}
MABs have also been widely proposed for dose-finding clinical trials, especially for  trials with a single drug \citep{aziz2019multi,shen2020icml,varatharajah2018contextual,villar2018covariate,lee2020contextual}.
{
Nevertheless, they do not address the challenging issues of Phase I clinical trials for multiple drugs, which operate in the {small sample size} regime relative to the number of potential doses. In trials for drug combination with more than ten potential DCs, their \textit{asymptotic} optimality is not meaningful for a practical number (fewer than 100) of patients in Phase I  trials.}
In MAB models, various safety management methods have been studied to address safety issue. However, they deal with the safety issue only implicitly or cannot be applied to the drug combination setting because they rely on a very limited dose-toxicity model with a single drug.\looseness=-1
\end{paragraph}

\begin{paragraph}{Methods for heterogeneous groups}
For both Bayesian and MAB methods, there has been only limited work involving heterogeneous groups.
In \cite{wages2015Phase}, a well-known Bayesian continual reassessment method (CRM) is extended for an adaptive clinical trial design for heterogeneous groups.
However, it does not consider drug combinations and a patient recruitment with a limited number of patients.
In \cite{atan2019sequential}, a patient recruitment with a limited number of patients is adaptively determined based on a Bayesian knowledge gradient policy, but its goal is to label the effectiveness of drugs as opposed to dose-finding.
\citep{varatharajah2018contextual,lee2020contextual} adapt a contextual MAB  for clinical trials with heterogeneous groups  by treating groups as contexts.  However, this work does not consider either drug combinations or patient recruitment with a limited number of patients.
\end{paragraph}

Table \ref{table:relatedwork} provides a summary comparison of our work with other methodologies.

\section{CONCLUSION}
In this paper, we have studied the problem of designing Phase I clinical trials for drug combinations.  We have enunciated a principle of cautious optimism and proposed the  SDF-Bayes algorithm that applies that principle to effectively balance the trade-off between the exploration of drug combinations and the risk of safety violation.  For settings with identified heterogeneous groups, we proposed an extension  SDF-Bayes-AR  in which both the DC to be allocated and the  group  from which the next patient is to be recruited are  chosen adaptively.  On the basis of experiments, we demonstrated that  our proposed algorithms outperform previous state-of-the-art algorithms.

\bibliography{mybib}

\onecolumn

\thispagestyle{empty}
\hsize\textwidth
  \linewidth\hsize \toptitlebar {\centering
  {\Large\bfseries Supplementary Material for SDF-Bayes: Cautious Optimism in  Safe Dose-Finding Clinical Trials with Drug Combinations and Heterogeneous Patient Groups \par}}
 \bottomtitlebar

\renewcommand\thesection{\Alph{section}}
\setcounter{section}{0}

\section{Joint Dose-Toxicity Models for Drug Combinations}
To reduce the search space in dose-finding for drug combinations,
various joint dose-toxicity models have been proposed.
They help us to efficiently investigate the toxicities of combinations of drugs.
We denote the vector of the parameters of joint-toxicity model as $\btheta=\{\balpha,\bbeta,\bgamma\}$, where $\balpha$ is a vector of the parameters that represents the relation between the toxicity effect and the dosage of drug A,
$\bbeta$ is a vector of the parameters that represents the relation between the toxicity effect and the dosage of drug B, and $\bgamma$ is a vector of the parameters that represents the relation between the toxicity effect and both dosages (i.e., drug-drug interaction).
The type of functions suitable for a joint dose-toxicity model is defined as the following admissibility conditions \citep{gasparini2013general}:
\begin{enumerate}
	\item $\pi(j,k,\btheta)$ is increasing separately in both $j$ and $k$
	\item There are functions $\pi_1(j,\balpha)$ and $\pi_2(k,\bbeta)$, which are called marginal dose-toxicity model such that $\pi(j,0,\btheta)=\pi_1(j,\balpha)$ and $\pi(0,k,\btheta)=\pi_2(k,\bbeta)$
	\item $\pi_1(0,\balpha)=\pi_2(0,\bbeta)=0$
\end{enumerate}


\begin{wrapfigure}{R}{0.4\textwidth}
	\center{
		\subfigure[Synergistic.]
		{
			\includegraphics[width=0.45\linewidth]{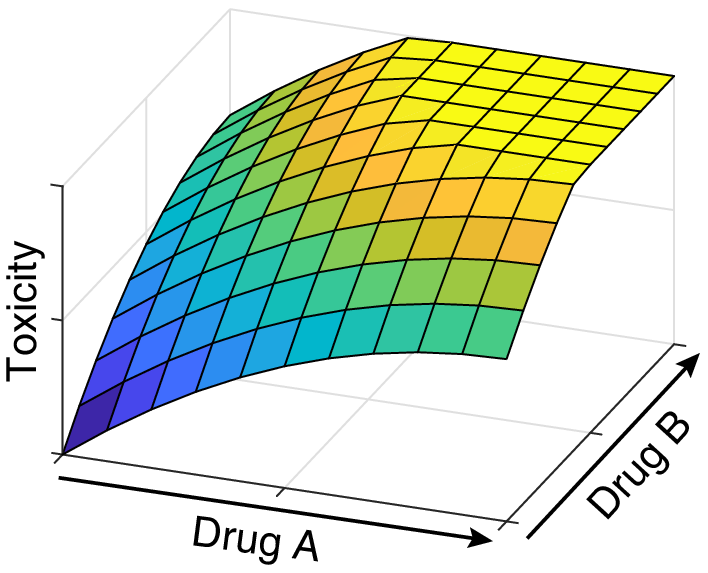}
			\label{fig:interaction_syn}
		}
		\subfigure[Antagonistic.]
		{
			\includegraphics[width=0.45\linewidth]{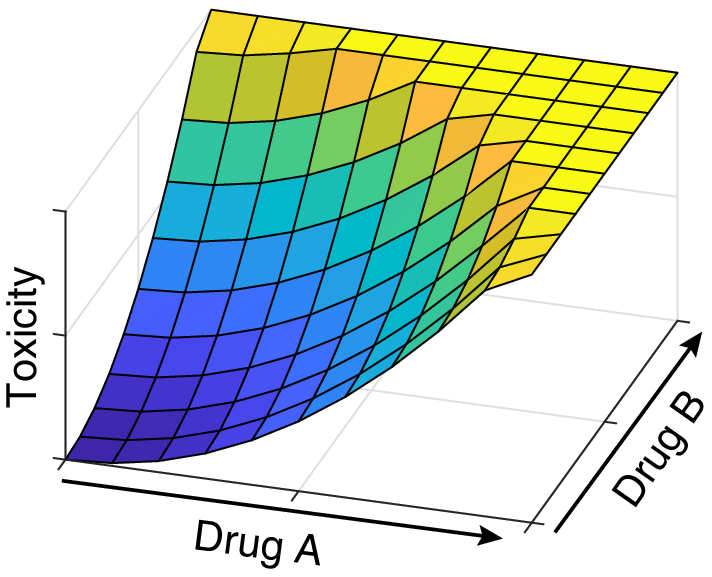}
			\label{fig:interaction_ant}
	}}\vspace{-1em}
	\caption{Examples of drug-drug interactions.\label{fig:interaction}}
\end{wrapfigure}

In joint dose-toxicity models, the drug-drug interaction such as synergy and antagonism shown in Fig. \ref{fig:interaction} should be captured.
To this end, various joint dose-toxicity models have been proposed in the literature as in Table \ref{table:dose-toxicity}.
In some models, the joint dose-toxicity model is defined by using the marginal model (i.e., $\pi_1(j,\balpha)$ and $\pi_2(k,\bbeta)$).
In this case, any dose-toxicity model for a single drug such as two-parameter logistic model can be used.
In the logistic model, the standardized dosage levels are used (i.e., $d_j$ and $u_k$ do not indicate actual dosage levels).
Thus, there is a parameter $\zeta$ that is not related to any drugs to ensure the appropriate toxicity modeling \citep{riviere2014bayesian}.
All the models except for the no interaction model can capture the drug-drug interaction and the parameter $\bgamma$ implies the characteristics of the interaction.
For the detailed description and discussion on drug-drug interactions, we refer the readers to \citep{gasparini2013general}.

\begin{table}[h]
	\caption{Summary of joint dose-toxicity models}
	\label{table:dose-toxicity}
	\begin{center}
		\begin{small}
			\setlength\tabcolsep{3pt}
			\begin{tabular}{cc}
				\toprule
				Model & $\pi(j,k,\btheta)$ \\
				\midrule
				No interaction & $1-\{1-\pi_1(j,\balpha)\}\{1-\pi_2(k,\bbeta)\}$ \\
				\midrule
				Constant log-odds difference & $\frac{1}{1+\gamma^{-1}[\{\pi_1(j,\balpha)+\pi_2(k,\bbeta)-\pi_1(j,\balpha)\pi_2(k,\bbeta)\}^{-1}-1]}$ \\
				\midrule
				Copula-based \citep{yin2009bayesian} & $1-C\{1-\pi_1(j,\balpha),1-\pi_2(k,\bbeta),\gamma\}$ \\
				\midrule
				Thall \citep{thall2003dose} & $\frac{\alpha_1 d_j^{\alpha_2}+\beta_1 u_k^{\beta_2}+\gamma\alpha_1 d_j^{\alpha_2}\beta_1 u_k^{\beta_2}}{1+\alpha_1 d_j^{\alpha_2}+\beta_1 u_k^{\beta_2}+\gamma\alpha_1 d_j^{\alpha_2}\beta_1 u_k^{\beta_2}}$ \\
				\midrule
				Exponential & $1-\exp\{ -(\alpha d_j+\beta u_k+\alpha\beta\gamma d_j u_k) \}$ \\
				\midrule
				Logistic \citep{riviere2014bayesian} & $\frac{1}{1-\exp(-\zeta-\alpha d_j-\beta u_k -\gamma d_j u_k)}$ \\
				\bottomrule
			\end{tabular}
		\end{small}
	\end{center}
\end{table}

\section{Proof of Proposition 1}
From the definition of $F_a^{\mathcal{O}}(v)$, in round $t$, we have
\[
\mathbb{P}\left[\left. p_a(\btheta)\leq F_a^{\mathcal{O}(t)}(v)\right|\mathcal{O}(t) \right]= v.
\]
Note that the event $\bigcap_{\tau=1}^t \left\lbrace \left.p_{a(\tau)}(\btheta)\leq F_{a(\tau)}(\btheta)^{\mathcal{O}(t)}(v)\right|\mathcal{O}(t) \right\rbrace$ is a subset of
the event $\left\lbrace\left.\sum_{\tau=1}^t p_{a(\tau)}(\btheta)\leq \sum_{\tau=1}^t F_a^{\mathcal{O}(t)}(v)\right|\mathcal{O}(t)\right\rbrace$ clearly.
Then, with $v=(1-\delta)^{1/t}$, we have
\[
\mathbb{P}\left[\left. \sum_{\tau=1}^t p_{a(\tau)}(\btheta)\leq \sum_{\tau=1}^t F_{a(\tau)}^{\mathcal{O}(t)}(v)\right|\mathcal{O}(t) \right] \geq \prod_{\tau=1}^t \mathbb{P}\left[\left.p_{a(\tau)}(\btheta)\leq F_{a(\tau)}^{\mathcal{O}(t)}(v)\right|\mathcal{O}(t) \right] = v^t = 1-\delta.
\]
From the proposition, the residual is non-negative, which implies that $(\xi+\epsilon_s)t\geq \sum_{\tau=1}^t F_a^{\mathcal{O}(t)}(v)$.
We then have
\[
\mathbb{P}\left[\left. \sum_{\tau=1}^t p_{a(\tau)}(\btheta)\leq (\xi+\epsilon_s)t\right|\mathcal{O}(t) \right]\geq \mathbb{P}\left[\left. \sum_{\tau=1}^t p_{a(\tau)}(\btheta)\leq \sum_{\tau=1}^t F_{a(\tau)}^{\mathcal{O}(t)}(v)\right|\mathcal{O}(t) \right] \geq 1-\delta.
\]

\section{StructMAB: Safe DC-Finding Bandits Based on Dose-Toxicity Structure}
\subsection{DC-Finding Bandits Based on Structured Bandits}

In our manuscript, we propose a DC-finding algorithm which allows the agent to effectively find the best recommendation for the MTD.
To this end, our algorithm exploits a dose-toxicity structure for the drugs from the joint dose-toxicity model.
Moreover, it is able to exploit arbitrary joint dose-toxicity models to ensure its practical use.
Since the conventional MAB-based dose-finding clinical trial algorithm in \cite{aziz2019multi} does not consider the dose-toxicity structure for drug combinations, we develop an advanced MAB-based algorithm by using structured bandits \citep{gupta2019unified}.

We first define a confidence set for the parameters as
\begin{equation}
	\label{eqn:conf_set_param}
	\hat{\Theta}(t)\coloneqq\left\lbrace \btheta: \forall a\in\mathcal{C},~|\bar{p}_{a}(t)-p_{a}(\btheta)|< \sqrt{\frac{\alpha\log t}{2n_a(t)}} \right\rbrace,
\end{equation}
where $\bar{p}_{a}(t)$ is the empirical toxicity of DC $a$ in round $t$ and $n_{a}(t)$ is the number of observed samples for DC $a$ until round $t$.
Since there is no assumption on the parameters $\btheta$, this confidence set can be obtained for arbitrary joint dose-toxicity models.
Then, by using the confidence set, we define a set of candidate MTDs as
\begin{equation}
	\label{eqn:set_candidate}
	\mathcal{A}(t)\coloneqq\left\lbrace a\in\mathcal{C} : \argmin_{a'\in\mathcal{C}}|p_{a'}(\btheta)-\xi|\textrm{ for some }\btheta\in\hat{\Theta}(t) \right\rbrace.
\end{equation}

For the candidate MTDs, we choose the DC by using Thompson sampling.
To this end, we assume that the expected toxicity of the DCs, $p_a$'s, are random variables.
The posterior probability of the expected toxicity of DC $a$ in round $t$ is updated as $\textrm{Beta}(s_a(t)+1,n_a(t)-s_a(t)+1)$.
Then, Thompson sampling is conducted on the set of candidate MTDs as
\begin{equation}
	\label{eqn:thompson_sampling}
	\tilde{p}_a(t)\sim\textrm{Beta}(s_a(t)+1,n_a(t)-s_a(t)+1),~\forall a\in\mathcal{A}(t).
\end{equation}
Based on the samples, the DC whose the sample is closest to the threshold is chosen as
\begin{equation}
	\label{eqn:chosen_DC}
	a(t)=\argmin_{a\in\mathcal{A}(t)}|\tilde{p}_a(t)-\xi|.
\end{equation}
At the end of the algorithm, the agent chooses the dose whose empirical toxicity is closest to $\xi$ as the MTD
\begin{equation}
	\label{eqn:recommended_DC}
	\hat{a}^*(T)=\argmin_{a\in\mathcal{C}}|\bar{p}_a(T)-\xi|
\end{equation}
or pick $\hat{a}^*$ uniformly at random among the allocated doses.

\subsection{Safe DC-Finding Bandits}
In clinical trials, it is important to avoid testing the unsafe DCs due to the ethical issues.
This can be achieved by considering the safety constraint in (1).
Here, we propose a safe DC-finding approach that satisfies the safety constraint as in the cautiousness of SDF-Bayes.

For the safe DC-finding approach, we first define a set of conservative DCs as
\begin{equation}
	\label{eqn:set_conservative}
	\mathcal{A}_c(t)\coloneqq \left\lbrace a\in\mathcal{C} : \max_{\btheta\in\hat{\Theta}(t)} p_a(\btheta)\leq \xi \right\rbrace.
\end{equation}
The DCs in the set of conservative DCs are safe (i.e., toxicity probability does not exceed the MTD threshold) with a high probability.
Thus, by choosing the DCs in the set only, we can avoid the trials with unsafe doses with a high probability.
However, this is too conservative and we cannot expect a good MTD recommendation.

To resolve this issue, we consider a residual for the safety constraint in each round.
The residual for the safety constraint in round $t$ is defined as
\[
r(t)=(\xi+\epsilon_s)t - \sum_{\tau=1}^{t-1} \max_{\btheta\in\hat{\Theta}(t)} p_{a(\tau)}(\btheta).
\]
Note that the confidence set for the parameters in round $t$, $\hat{\Theta}(t)$, is used for the calculation of the sum of the expected toxicities until round $t$.
By using the residual, we can infer whether the safety constraint will be violated or not based on the current confidence set for the parameters.
If we still have the non-negative residual with the chosen DC in \eqref{eqn:chosen_DC} (i.e., $r(t)-\max_{\btheta\in\hat{\Theta}(t)}p_{a(t)}(\btheta)\geq 0$), then we can expect the safety constraint not to be violated with the chosen DC.
Thus, in that case, the safe DC-finding approach accepts the chosen DC, and otherwise, it rejects the chosen DC and chooses the DC in the conservative set $\mathcal{A}_c(t)$ instead to ensure the positive residual.
We summarize the safe DC-finding algorithm (StructMAB) in Algorithm \ref{alg:safe_dc_finding}.

\begin{algorithm}[tb]
	\caption{\textsc{StructMAB}}
	\label{alg:safe_dc_finding}
	\begin{algorithmic}[1]
		\While{$t\leq T$}
		\State Obtain a confidence set $\hat{\Theta}(t)$ as in \eqref{eqn:conf_set_param}
		\State Obtain a set of candidate MTDs $\mathcal{A}(t)$ as in \eqref{eqn:set_candidate}
		\State Sample the expected toxicities for all candidate MTDs as in \eqref{eqn:thompson_sampling}
		\State $\bar{a}\leftarrow \argmin_{a\in\mathcal{A}(t)}|\tilde{p}_a(t)-\xi|$
		\If{$r(t)-\max_{\btheta\in\hat{\Theta}(t)}p_{\bar{a}}\geq 0$}
		\State $a(t)\leftarrow \bar{a}$
		\Else
		\State Obtain a set of conservative DCs $\mathcal{A}_c(t)$ as in \eqref{eqn:set_conservative}
		\State Sample the expected toxicities for all conservative DCs
		\State $a(t)\leftarrow \argmin_{a\in\mathcal{A}_c(t)}|\tilde{p}_a(t)-\xi|$
		\EndIf
		\State Observe the DLT $Y_t$
		\State Update $m_{a(t)}(t+1)$ and $n_{a(t)}(t+1)$
		\State $t\leftarrow t+1$
		\EndWhile
	\end{algorithmic}
\end{algorithm}

%
%

\subsection{Sampling-Based Implementation of StructMAB}
To use StructMAB in practice, we need to identify the confidence set $\hat{\Theta}$ and the set of candidate MTDs $\mathcal{A}$ based on the confidence set, which requires a high computational complexity in general.
Besides, we need different identification methods for different joint dose-toxicity model used in the algorithm.
Thus, to resolve these issues, we propose a simple sampling method for StructMAB that can efficiently approximate the confidence set and set of candidate MTDs regardless of which joint dose-toxicity model is used.


Here, we sample $\btheta$ from the posterior distribution of $\btheta$, $f(\btheta|\mathcal{O})$, by using Gibbs sampling which is one of most representative Bayesian sampling algorithms for multidimensional sampling as in SDF-Bayes.
The posterior distribution of $\btheta$ can be found in Section 3.1. of our manuscript.
Details of the Gibbs sampling procedure can be found in the following section.
We denote the number of samples from the posterior distribution $p(\btheta|\mathcal{O}(t))$ by $L$ and the samples by $\tilde{\Theta}(t)=\{\btheta^{(l)}\}_{l\in[L]}$.
Then, from the samples, we can define an approximated confidence set for the parameters as
\[
\hat{\Theta}'(t)\coloneqq \left\lbrace \btheta^{(l)}\in\tilde{\Theta}(t) : \forall a\in\mathcal{C},~|\bar{p}_{a}(t)-p_{a}(\btheta^{(l)})|< \sqrt{\frac{\alpha\log t}{2n_a(t)}} \right\rbrace
\]
and an approximated set of candidate MTDs as
\begin{equation}
	\label{eqn:approx_candidate_set}
	\mathcal{A}'(t)\coloneqq\left\lbrace a\in\mathcal{C} : \argmin_{a'\in\mathcal{C}}|p_{a'}(\btheta^{(l)})-\xi|\textrm{ for some }\btheta^{(l)}\in\hat{\Theta}'(t) \right\rbrace.
\end{equation}
Similarly, we can define an approximated conservative DCs as
\[
\mathcal{A}_c'(t)\coloneqq\left\lbrace a\in\mathcal{C} : p_{a'}(\btheta^{(l)})\leq \xi,~\forall\btheta^{(l)}\in\hat{\Theta}'(t) \right\rbrace.
\]
We can use these approximated sets for StructMAB.
Note that these approximations converges to the true sets
as the number of samples goes to infinity.
Moreover, for the residual, we use $\max_{\btheta^{(l)}\in\tilde{\Theta}(t)}p_{a(t)}(\btheta^{(l)})$.

\section{Description of Sampling Procedure}
We describe the sampling procedure used in SDF-Bayes.
\begin{itemize}
	\item
	\textbf{Gibbs sampling}: Gibbs sampling is one of the most representative Markov chain Monte Carlo (MCMC) sampling methods to generate a sequence of samples for multiple variables from a multivariate joint probability distribution.
	In SDF-Bayes, it is used to generate a sequence of samples of the parameter vector $\btheta=\{\theta_0,\theta_1,...,\theta_{D_\btheta-1}\}$ from $f(\btheta|\mathcal{O})$, where $D_{\btheta}$ is the dimension of the parameter vector.
	Note that the dimension of the parameter vector depends on which joint toxicity model is used for SDF-Bayes.
	In Gibbs sampling, we start from an arbitrary initial sample $\btheta$ in the distribution.
	We then samples each component of the parameter vector in turn based on each of the full conditional distribution with updated parameter samples.
	In Gibbs sampling, we discard $L_b$ samples at the beginning which is so-called burn-in period, and then, retain the following $L$ samples.
	We formally summarize Gibbs sampling as following Algorithm \ref{alg:Gibbs_sampling}.

	\begin{algorithm}[h]
		\caption{\textsc{Gibbs sampling}}
		\label{alg:Gibbs_sampling}
		\begin{algorithmic}[1]
			\State Initialize $\btheta^{(0)}$
			\For{$l=1$ {\bfseries to} $L_b+L$}
			\For{$d=0$ {\bfseries to} $D_{\btheta}-1$}
			\State Sample $\theta_d^{(l)}$ from its full conditional distribution $f\left(\theta_d\left|\theta_0^{(l)},...,\theta_{d-1}^{(l)},\theta_{d+1}^{(l-1)},...,\theta_{D_{\btheta}-1}^{(l-1)},\mathcal{O}\right.\right)$
			\EndFor
			\EndFor
		\end{algorithmic}
	\end{algorithm}

	\item \textbf{Adaptive rejection metropolis sampling}:
	Gibbs sampling relies on the complete full conditional distributions of all components.
	However, in general, we cannot access to closed forms of such distributions.
	Thus, in SDF-Bayes, we use adaptive rejection metropolis sampling (ARMS) within Gibbs sampling \citep{gilks1995adaptive}.
	ARMS is a MCMC sampling method as well, and it used to sample from a univariate target distribution specified by (unnormalized) log density.

	Thus, we utilize ARMS to sample each component of the parameter vector from its univariate full conditional distribution (i.e., for line 4 of Algorithm \ref{alg:Gibbs_sampling}).
	We assume that the prior distributions of the parameters are independent.
	We denote the parameter vector except for $d$-th component by $\btheta_{-d}$.
	Then, we can obtain the unnormalized density of the full conditional distribution of $d$-th parameter, $f(\theta_d|\btheta_{-d}',\mathcal{O})$, from the likelihood of $\btheta$ and the prior distribution as
	\[
	f(\theta_d|\btheta_{-d}',\mathcal{O})\propto L(\theta_d,\btheta_{-d}'|\mathcal{O})f(\theta_d)\prod_{d'\in\{1,...,d-1,d+1,...D_{\btheta}\}}f(\theta_{d'}').
	\]
	By using this density, we can sample each component of the parameters within Gibbs sampling.
	For the detail procedure of ARMS, we refer to \citep{gilks1995adaptive}.
\end{itemize}

\section{Description of SDF-Bayes-AR}

\begin{algorithm}[H]
	\caption{\textsc{SDF-Bayes-AR}}
	\label{alg:safe_dc_finding_bayes_group}
	\begin{algorithmic}[1]
		\While{$t\leq T$}
		\State Obtain $a_m(t)$'s by \textsc{SDF-Bayes} (lines 2--13)
		\State $g(t)\leftarrow\argmax_{m\in\mathcal{M}}\tilde{H}^{\mathcal{O}(t)}_{m,a_m(t)}(u)$
		\State $a(t)\leftarrow a_{g(t)}(t)$
		\State Observe DLT $Y_t$
		\State Update $s^{g(t)}_{a(t)}(t+1)$ and $n^{g(t)}_{a(t)}(t+1)$
		\State $t\leftarrow t+1$
		\EndWhile
		\State \textbf{Output:} $\hat{a}^*_m\!=\!\argmax_{a\in\mathcal{A}}\tilde{G}_{m,a}^{\mathcal{O}(T)}(u),~\forall m\!\in\!\mathcal{M}$
	\end{algorithmic}
\end{algorithm}

\section{Description of Detailed Experiment Settings}

\subsection{Real-World Dataset}
We describe the real-world dataset used in the experiments.
We consider a Phase I drug combination clinical trial dataset and its corresponding dose-toxicity model provided in \cite{bailey2009bayesian}. In the dataset, the drug combination of nilotinib and imatinib is considered and the DLT observations from 50 patients are provided  to assess the toxicity of different dose combinations. Furthermore, in \cite{bailey2009bayesian}, the dose-toxicity model of the combination is constructed in a Bayesian way based on a logistic regression model by using the DLT observations and the prior information about the drugs. In specific, the model is given by
\[
\textrm{logit}(d,X_1,X_2,X_3)=\log(\alpha)+\beta\log(d/d^*)+\xi_1X_1+\xi_2X_2+\xi_3X_3,
\]
where $\alpha$, $\beta$, $\xi_n$'s are dose-toxicity model parameters, $d$ represents the doses of nilotinib and $d^*$ is the reference dose of nilotinib (400mg) for scaling. Also, $(X_1,X_2,X_3)$ represents the dose of imatinib by taking the form $(0,0,0)$, $(1,0,0)$, $(1,1,0)$, and $(1,1,1)$ for imatinib doses 0, 400, 600, and 800mg, respectively. The prior distributions of the parameters are tuned by using the historical data such as previous clinical data of each drug without combinations. We rounded up the toxicities at the third decimal place. 
This trial is originally designed to find the DCs whose toxicities belong to the target interval $(0.20,0.35)$.
This implies that the potential DCs of the trial are more conservative in terms of safety (toxicity) compared with typical synthetic datasets designed to find the DC whose toxicity is close to the target toxicity 0.3
because it should find the DCs whose toxicities is lower than 0.3 in average.


\subsection{Description of Algorithms in Experiments}
Here, we describe the algorithms used in the experiments. We set $u=0.1$ for all the Bayesian algorithms.
\begin{itemize}
	\item \textbf{SDF-Bayes}:
	We implement SDF-Bayes based on Algorithm 1 in our manuscript.
	Here, we provide the settings of SDF-Bayes in the experiments.
	We set $v=0.9$ unless mentioned.
	For warm-start of SDF-Bayes, the residual at the early stage of trials can be given by $r(t) = \max\left( (\xi+\epsilon_s)t-\sum_{\tau=1}^{t-1}\tilde{F}_{a(\tau)}^{\mathcal{O}(t)}(v), R \right)$, where $R$ is a constant.
	Typically, $\xi T$ works well for the cases with a single group.
	For the joint dose-toxicity model, we use the following logistic dose-toxicity model proposed in \cite{riviere2014bayesian}:
	\[
	\texttt{logit}(\pi_{jk})=\theta_0+\theta_1 u_j+\theta_2 v_k + \theta_3 u_j v_k,
	\]
	where $\theta_0$, $\theta_1$, $\theta_2$, and $\theta_3$ are parameters and $u_j$ and $v_k$ are the standardized dose of drugs.
	In the literature \citep{riviere2014bayesian}, $u_j$'s and $v_k$'s are defined as $u_j=\log\left(\frac{p_j}{1-p_j}\right)$ and $v_k=\log\left(\frac{q_k}{1-q_k}\right)$, where $p_j$ and $q_k$ are the prior estimates of the toxicity probabilities of the $j$-th dosage of drug A and $k$-th dosage of drug B.
	However, such prior information is not always available, and thus, in the algorithm, we simply use $(-2,-1,0)$ for $u_j$'s and $(-3,-2,-1,0)$ for $v_k$'s assuming without any prior information.
	In this model, the parameters are defined as $\theta_1>0$ and $\theta_2>0$ and $-\infty<\theta_0<\infty$.
	In addition, $\theta_3$ should satisfy $\theta_1+\theta_3 v_k>0$ and $\theta_2+\theta_3 u_j$ for all $k\in\mathcal{K}$ and $j\in\mathcal{J}$.
	This ensures that the increasing toxicity probability with the increasing dose of a single drug.
	For the prior of the parameters, we use a normal distribution $\mathcal{N}(0,10)$ for $\theta_0$ and $\theta_3$.
	This prior is vague with relatively high variance.
	For $\theta_1$ and $\theta_2$, we use an exponential distribution $\textrm{Exp}(1)$ since they are positive.
	These default prior settings are same with used in the literature \citep{riviere2014bayesian} and used for the other algorithms with the joint dose-toxicity model.
	For dataset RW, we use $v=0.85$ because the potential DCs of the dataset are designed to be more conservative in terms of safety compared with other datasets as described in the previous section.

	\item \textbf{SOTA Bayes} \citep{riviere2014bayesian}:
	We implement a Bayesian dose-finding algorithm in \cite{riviere2014bayesian}.
	We denote the probability threshold for dose escalation by $c_e$ and the probability threshold for dose de-escalation by $c_d$.
	To avoid that the dose is determined to be escalated and de-escalated at the same time, we need to ensure $c_e+c_d>1$.
	We use the joint dose-toxicity model with the same setting from SDF-Bayes for fair comparison.
	The dose-finding algorithm proposed in \cite{riviere2014bayesian} is described in following:
	\begin{itemize}
		\item \textit{Dose escalation}:
		Let the current DC be $(j,k)$.
		If $\mathbb{P}[p_{jk}<\xi|\mathcal{O}]>c_e$, then the current DC is escalated to an adjacent DC among $\{(j+1,k),(j,k+1),(j+1,k-1),(j-1,k+1)\}$ that has a toxicity probability higher than $p_{jk}$ and closest to $\xi$.
		If the current DC is the highest one, the same DC is allocated.
		With the samples from the posterior distribution $f(\btheta|\mathcal{O})$, we approximate the condition as $\frac{1}{L}\sum_{l=1}^L \mathbb{I}[p_{jk}(\btheta^{(l)})<\xi]>c_e$.

		\item \textit{Dose de-escalation}:
		Let the current DC be $(j,k)$.
		If $\mathbb{P}[p_{jk}>\xi|\mathcal{O}]>c_d$, then the current DC is de-escalated to an adjacent DC among $\{(j-1,k),(j,k-1),(j+1,k-1),(j-1,k+1)\}$ that has a toxicity probability lower than $p_{jk}$ and closest to $\xi$.
		If the current DC is the lowest one, the same DC is allocated.
		With the samples from the posterior distribution $f(\btheta|\mathcal{O})$, we approximate the condition as $\frac{1}{L}\sum_{l=1}^L \mathbb{I}[p_{jk}(\btheta^{(l)})>\xi]>c_d$.

		\item \textit{Dose retainment}:
		Let the current DC be $(j,k)$.
		If $\mathbb{P}[p_{jk}<\xi|\mathcal{O}]\leq c_e$ and $\mathbb{P}[p_{jk}>\xi|\mathcal{O}]\leq c_d$, the current DC is allocated.
		With the samples from the posterior distribution $f(\btheta|\mathcal{O})$, we approximate the condition as $\frac{1}{L}\sum_{l=1}^L \mathbb{I}[p_{jk}(\btheta^{(l)})<\xi]\leq c_e$ and $\frac{1}{L}\sum_{l=1}^L \mathbb{I}[p_{jk}(\btheta^{(l)})>\xi]\leq c_d$.
	\end{itemize}

	At the end of the trial, the DC recommendation is determined as same with SDF-Bayes.

	\item \textbf{StructMAB}:
	We implement StructMAB based on Algorithm \ref{alg:safe_dc_finding}.
	For the algorithm, we use the same joint dose-toxicity model and settings for the model used in SDF-Bayes.
	We set the exploration parameter $\alpha$ in the confidence set for the parameters in \eqref{eqn:conf_set_param} to be $1$.
	Moreover, when constructing the set of candidate MTDs in \eqref{eqn:approx_candidate_set}, its condition to include a DC into the set of candidate MTDs is too sensitive in practice since a DC will be included into the set even with only one ground sample that asserts that the DC is optimal.
	Thus, to mitigate such sensitivity of the condition, we add the following condition: from the set, remove the DCs whose number of ground samples is below the 20\% percentile of the number of ground samples of each DC.
	This helps to remove only the outliers from the set of candidate MTDs.
	For example, when all the DCs in the set have only one ground sample, then the condition does not remove any DC from the set.
	Similarly, to avoid StructMAB being too conservative, we use the 80\% percentile of the toxicity of the allocated DC among the samples when calculating the residual for the safety constraint.

	\item \textbf{IndepTS} \citep{aziz2019multi}:
	We extend an independent Thompson sampling algorithm in \cite{aziz2019multi} for drug combinations.
	To this end, we simply expand an arm space of IndepTS into two dimensional space for drug combinations.
	In IndepTS, Thompson sampling is used to estimate the toxicity of DC $a$ as follows:
	\[
	\tilde{p}_{a}(t)\sim \textrm{Beta}(\alpha_{a}(t),\beta_{a}(t)),
	\]
	where
	$\alpha_{a}(t)=s_{a}(t)+1$ and
	$\beta_a(t)=n_{a}(t)-s_a(t)+1$.
	In each round $t$, the expected toxicities of all DCs are sampled based on the above equation.
	Then, the DC $a(t)$ that has the maximum toxicity sample $\tilde{p}_a(t)$ is chosen (i.e., $a(t)=\argmin_{a\in\mathcal{C}}|\tilde{p}_a(t)-\xi|$).
	At the end of the trial, it recommends a DC as: $\hat{a}^*(T)=\argmin_{a\in\mathcal{C}}|\hat{p}_{s,k}(T)-\xi|$.
	Note that the implicit safety consideration in IndepTS is not exploited in IndepTS-DC since it is for a single drug.

	\makebox[\linewidth]{\rule{\linewidth}{0.4pt}}
	\item \textbf{SDF-Bayes-AR}:
	We implement SDF-Bayes-AR based on Algorithm \ref{alg:safe_dc_finding_bayes_group} in this supplementary material.
	In a practical implementation of SDF-Bayes-AR, we adopt uniform patient recruitment in an early phase of clinical trial with the rounds $t\leq T/4$.
	This enables a warm-start of adaptive patient recruitment and avoids biased recruitment.
	In addition, we adopt an early stopping strategy to adaptive patient recruitment that has been widely considered in clinical literatures \citep{riviere2014bayesian}.
	If the posterior probability of the most likely DC to be the MTD for group $m$ exceeds a threshold (i.e., $\max_{a\in\mathcal{C}}\tilde{G}_{m,a}^{\mathcal{O}(t)}(u)>p_{es}$, where $p_es$ is the threshold), then SDF-Bayes-AR stops recruiting the patients from group $m$.
	This prevents unnecessary patient recruitment for the groups whose MTD is confidently discovered.

	\item \textbf{SOTA Bayes-AR}: We implement SOTA Bayes-AR by adopting our proposed adaptive patient recruitment (AR) to SOTA Bayes. For this, we can simply change line 2 in Algorithm \ref{alg:safe_dc_finding_bayes_group} as ``Obtain $a_m(t)$'s by SOTA Bayes''.\looseness=-1

\end{itemize}

\subsection{Experiments with Heterogeneous Groups}

\begin{wraptable}{R}{4.4cm}
\vspace{-0.3in}
\fontsize{9pt}{9pt}\selectfont
\centering
\caption{True toxicity of EP \label{table:toxicity_ep}}
\vspace{0.1in}
\setlength\tabcolsep{1.5pt}
\begin{tabular}{c|cccc}
	\toprule
	&       \multicolumn{4}{c}{Synthetic EP}        \\ \midrule
	~~~~~ &   1   &   2   &       3       &       4       \\ \midrule
	3   & 0.12  & 0.21  & \textbf{0.30} &     0.40      \\
	2   & 0.075 & 0.125 &     0.215     & \textbf{0.30} \\
	1   & 0.035 & 0.09  &     0.125     &     0.205     \\ \bottomrule
\end{tabular}
\end{wraptable}

In the experiments with heterogeneous groups, we apply the Bayesian algorithms to the entire population by treating it as a single group called EP. The toxicity probabilities for EP are the averages of the toxicity probabilities for group A and B, which are provided in Table \ref{table:toxicity_ep}.
In the results, the safety violations of each group and the entire trial are considered. For each simulated trial, the safety violation of each group occurs if $S_m(T)> \xi+\epsilon$ as described in Section 4.1. On the other hand, the safety violation of the entire trial occurs if $\frac{1}{T}\sum_{t=1}^T Y_t > \xi+\epsilon$.
For prior information, we run a simulated trial for group B with $T_p$ patients using SDF-Bayes. Then, we use the observations from the simulated trial as the prior information.

\section{More Experiment Results with Homogeneous Group}

\subsection{DLT Observation Rates with Datasets and Distribution of DLT Observation Rates}

\begin{table}[h]
	\fontsize{9pt}{9pt}\selectfont
	\caption{DLT observation rates with datasets A,B,C,D,RW.}
	\label{table:dlt_observation}
	\centering
	\vspace{0.1in}
	\setlength\tabcolsep{1.5pt}
	\begin{tabular}{c||c|c|c|c|c}
		\toprule
		           &  Synthetic A   &  Synthetic B   &  Synthetic C   &  Synthetic D   &   Real-World    \\ \midrule
		SDF-Bayes  & 0.296$\pm$.001 & 0.245$\pm$.001 & 0.286$\pm$.001 & 0.305$\pm$.001 & 0.274$\pm$.001  \\
		 DF-Bayes  & 0.342$\pm$.002 & 0.260$\pm$.001 & 0.343$\pm$.001 & 0.351$\pm$.001 & 0.303$\pm$.001  \\
		SOTA Bayes & 0.268$\pm$.001 & 0.204$\pm$.001 & 0.268$\pm$.001 & 0.275$\pm$.001 & 0.238$\pm$.001  \\
		StructMAB  & 0.282$\pm$.001 & 0.184$\pm$.001 & 0.279$\pm$.001 & 0.301$\pm$.001 & 0.238$\pm$.001  \\
		 IndepTS   & 0.239$\pm$.001 & 0.118$\pm$.001 & 0.233$\pm$.001 & 0.278$\pm$.001 & 0.176$\pm$.001  \\ \bottomrule
	\end{tabular}
\end{table}

From Table \ref{table:dlt_observation}, SDF-Bayes satisfies the safety constraint but it does not achieve the lowest DLT observation rate because it is more often testing DC's that are believed close to being unsafe; this allows it to more effectively find the true MTD. This exploration-toxicity trade-off can be seen even more clearly by looking at the distribution of DLT observation rates in the following.

\begin{figure}[h]
	\begin{center}
		\includegraphics[width=1.\textwidth]{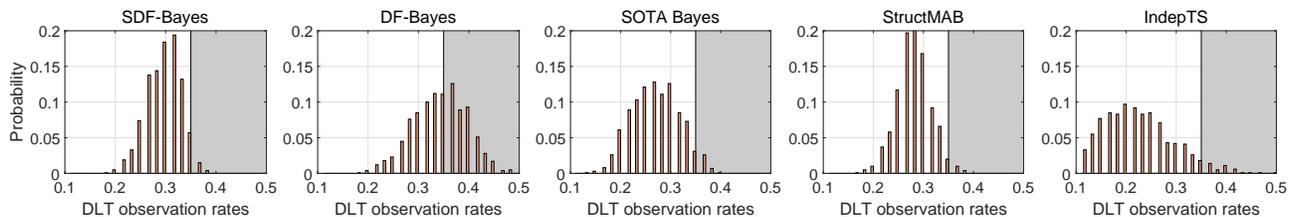}
		\caption{The distribution of DLT observation rates. The region shaded in gray is the region of the DLT observation rates that violate the safety constraint.}
		\label{fig:dlt_dist}
	\end{center}
\end{figure}

In Fig. \ref{fig:dlt_dist}, we provide the histogram of the DLT observation rates of the trials with synthetic dataset A.
From the figure, we can see that the DLT observation rates with SDF-Bayes are distributed close to the safety threshold, but only a few trials violate the safety constraint.
This clearly shows that SDF-Bayes effectively balances the trade-off between the exploration of the toxicities of the DCs and the DLT observation as intended in its cautious optimism.
On the other hand, DF-Bayes frequently violates the safety constraint.
This shows the effectiveness of the risk management by the cautiousness in SDF-Bayes.
The safety constraint is rarely violated with SOTA Bayes as much as SDF-Bayes.
However, their DLT observation rates do not concentrate as in SDF-Bayes since its dose escalation does not have a capability to balance the trade-off explicitly.
StructMAB has the similar distribution of DLT observation rates with SDF-Bayes, since the cautiousness principle of SDF-Bayes is adopted in StructMAB.
The DLT observation rates of IndepTS are distributed over a low rate region since it fails to estimate the toxicities of DCs.

\subsection{DC Allocation Ratios}

\begin{figure}[h]
	\begin{center}
		\includegraphics[width=0.55\columnwidth]{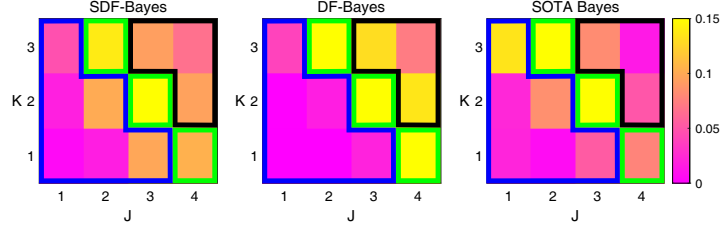}
		\vspace{-0.1in}
		\caption{Heatmaps for DC allocation ratios of Bayesian algorithms. The regions with blue, green, and black borders are underdosing, target-dose, and overdosing regions, respectively.}
		\label{fig:sel_heatmap}
	\end{center}
\end{figure}

Fig. \ref{fig:sel_heatmap} shows  heatmaps for the DC allocation ratios of the Bayesian algorithms.  We can see that  the optimism of DF-Bayes results in allocating the most DCs to the target-dose and overdosing.
SOTA Bayes concentrates  DCs near the MTD $(3,2)$ because of the way it escalates doses.  On the other hand, SDF-Bayes allocates DCs in a less  concentrated fashion because it tempers its optimism with caution.  The result is that SDF-Bayes obtains more information than the other algorithms (and hence makes fewer recommendation errors) while managing risk better.\looseness=-1

\subsection{Results with More Toxicity Probability Models}

\begin{table}[h]
	\vspace{-0.2in}
	\caption{Synthetic models.}
	\label{table:synthetic_models}
	\begin{center}
		\begin{small}
			\setlength\tabcolsep{1.5pt}
			\begin{tabular}{c|cccc|cccc|cccc|cccc|cccc}
				\toprule
				      &      \multicolumn{4}{c|}{Synthetic E}       &       \multicolumn{4}{c|}{Synthetic F}       &           \multicolumn{4}{c|}{Synthetic G}            &       \multicolumn{4}{c|}{Synthetic H}       &           \multicolumn{4}{c}{Synthetic I}            \\ \midrule
				~~~~~ &  1   &       2       &  3   &       4       &  1   &  2   &       3       &       4       &  1   &       2       &       3       &       4       &  1   &  2   &       3       &       4       &  1   &       2       &       3       &       4       \\ \hline
				  3   & 0.15 & \textbf{0.30} & 0.45 &     0.50      & 0.10 & 0.15 & \textbf{0.35} &     0.50      & 0.17 & \textbf{0.35} &     0.45      &     0.50      & 0.10 & 0.15 & \textbf{0.25} &     0.40      & 0.10 & \textbf{0.25} &     0.40      &     0.60      \\
				  2   & 0.09 &     0.12      & 0.15 & \textbf{0.30} & 0.07 & 0.12 &     0.16      & \textbf{0.35} & 0.10 &     0.17      & \textbf{0.35} &     0.45      & 0.07 & 0.12 &     0.16      & \textbf{0.25} & 0.07 &     0.12      & \textbf{0.25} &     0.40      \\
				  1   & 0.05 &     0.08      & 0.10 &     0.13      & 0.03 & 0.06 &     0.08      &     0.10      & 0.05 &     0.10      &     0.17      & \textbf{0.35} & 0.03 & 0.06 &     0.08      &     0.10      & 0.03 &     0.08      &     0.18      & \textbf{0.25} \\ \bottomrule
			\end{tabular}
		\end{small}
	\end{center}
\end{table}

\begin{table}[h]
	\vspace{-0.1in}
	\fontsize{7pt}{7pt}\selectfont
	\caption{Safety constraint violation rates and DC recommendation error rates with different datasets.}
	\label{table:performances_models}
	\centering
	\vspace{0.1in}
	\setlength\tabcolsep{0.3pt}
	\begin{tabular}{c||c|c||c|c||c|c||c|c||c|c}
		\toprule
		           &       \multicolumn{2}{c||}{Synthetic E}        &       \multicolumn{2}{c||}{Synthetic F}        &       \multicolumn{2}{c||}{Synthetic G}        &       \multicolumn{2}{c||}{Synthetic H}        &        \multicolumn{2}{c}{Synthetic I}         \\ \midrule
		Algorithms &     Safety vio.      &    \makecell{Errors}    &     Safety vio.      &    \makecell{Errors}    &     Safety vio.      &    \makecell{Errors}    &     Safety vio.      &    \makecell{Errors}    &     Safety vio.      &    \makecell{Errors}    \\ \midrule
		SDF-Bayes  &    0.010$\pm$.002    & \textbf{0.295$\pm$.011} &    0.006$\pm$.002    &     0.251$\pm$.010      &    0.035$\pm$.004    &     0.285$\pm$.011      &    0.003$\pm$.001    & \textbf{0.331$\pm$.011} &    0.033$\pm$.004    & \textbf{0.298$\pm$.011} \\
		 DF-Bayes  & \vio{0.239$\pm$.010} &     0.235$\pm$.010      & \vio{0.245$\pm$.010} &     0.195$\pm$.009      & \vio{0.530$\pm$.012} &     0.277$\pm$.010      & \vio{0.112$\pm$.007} &     0.359$\pm$.011      & \vio{0.300$\pm$.011} &     0.392$\pm$.011      \\
		SOTA Bayes &    0.005$\pm$.002    &     0.318$\pm$.011      &    0.005$\pm$.002    & \textbf{0.214$\pm$.010} &    0.049$\pm$.005    & \textbf{0.283$\pm$.010} &    0.001$\pm$.001    &     0.338$\pm$.011      &    0.014$\pm$.003    &     0.361$\pm$.011      \\
		StructMAB  &    0.001$\pm$.001    &     0.592$\pm$.011      &    0.001$\pm$.001    &     0.547$\pm$.012      &    0.033$\pm$.004    &     0.616$\pm$.011      &    0.000$\pm$.000    &     0.490$\pm$.012      &    0.010$\pm$.002    &     0.445$\pm$.012      \\
		 IndepTS   &    0.000$\pm$.000    &     0.726$\pm$.010      &    0.000$\pm$.000    &     0.697$\pm$.011      &    0.029$\pm$.004    &     0.632$\pm$.011      &    0.000$\pm$.000    &     0.712$\pm$.011      &    0.004$\pm$.001    &     0.582$\pm$.011      \\ \bottomrule
	\end{tabular}
\end{table}

Here, we introduce 5 more different synthetic datasets to consider various possible combination toxicities.
The toxicity probabilities of each synthetic model are summarized in Table \ref{table:synthetic_models} and the MTDs are highlighted in bold.
Datasets F and G describe the scenarios in which the MTD's toxicity probability is higher than the target toxicity ($\xi=0.3$) and datasets H and I describe the scenarios in which the MTD's toxicity probability is lower than the target toxicity.
In Table \ref{table:performances_models}, we provide the performance of the algorithms with the additional synthetic model in Table \ref{table:synthetic_models}.
From the results, we can see that in terms of recommendation error rates, SDF-Bayes outperforms SOTA Bayes in datasets E, H, and I.
On the other hand, in datasets F and G, SOTA Bayes achieves lower error rates.
This is because SDF-Bayes cautiously chooses the DC to be allocated considering the safety violation; it is hard to follow the optimism principle in those datasets because the MTDs have the higher toxicity than the target toxicity.
Hence, SDF-Bayes has an advantage in terms of safety as shown in the results of dataset G; SOTA Bayes is very close to the boundary of a failure to satisfy the safety constraint while SDF-Bayes is not.
In the results of MAB-based algorithms (i.e., StructMAB and IndepTS), StructMAB outperforms IndepTS in terms of error rates and achieves similar or lower safety violation rates.
However, their error rates are too high compared with those of the other Bayesian algorithms.

\clearpage
\subsection{Impact of Budgets}

Fig. \ref{fig:bandit_var_T} reports the error rate as a function of the total patient budget. We see that in the practical regime of budgets (less than 100), the error rates of the MAB-based algorithms are significantly higher than the Bayesian algorithms. We note that this is even by using {\em StructMAB}, which already exploits the dose-toxicity structure to speed up learning. This clearly shows that the Bayesian designs are more suitable than the MAB-based designs in practice in terms of error rates.\looseness=-1

\begin{figure}[h]
	\begin{center}
		\includegraphics[width=0.35\textwidth]{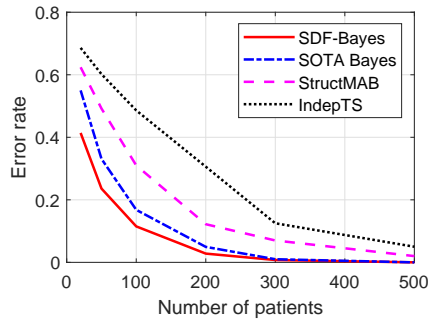}
		\caption{Error rates varying budgets. }
		\label{fig:bandit_var_T}
	\end{center}
	\vskip -0.2in
\end{figure}

\begin{wraptable}{R}{0.44\linewidth}
	\vspace{-0.2in}
	\caption{Performances of SDF-Bayes varying $v$}
	\vskip 0.1in
	\label{table:varying_v}
	\begin{center}
		\begin{small}
			\begin{tabular}{c||c|c|c}
				\toprule
				$v$ & \makecell{Safety vio.} & \makecell{Rec. errors} & \makecell{DLT observ.} \\
				\midrule
				0.80 & 0.135 & 0.191 & 0.313 \\
				0.85 & 0.056 & \textbf{0.188} & 0.313 \\
				0.90 & 0.022 & 0.203 & 0.296 \\
				0.95 & \textbf{0.005} & 0.222 & \textbf{0.279} \\
				\bottomrule
			\end{tabular}
		\end{small}
	\end{center}
	\vskip -0.1in
\end{wraptable}

\subsection{Impact of Hyperparameter $v$}
\label{sec:exp:hyperparameter}
To show the impact of hyperparameter $v$ on the performance of SDF-Bayes, we provide the performance of SDF-Bayes varying $v$ from 0.80 to 0.95 in Table \ref{table:varying_v}.
Note that the hyperparameter $v$ controls how conservative the algorithm is.
From the results, we can see that the recommendation error rate increases as $v$ increases.
On the other hand, the safety constraint violation rate decreases as $v$ increases.
These results clearly show that the hyperparameter $v$ controls the conservativeness of SDF-Bayes well.
When $v$ is increasing, SDF-Bayes chooses the DCs more conservatively.
Thus, the risk of safety constraint violation decreases.
On the other hand, it results in a less exploration.
Then, the recommendation error rate increases.

\begin{wraptable}{R}{0.4\linewidth}
	\vspace{-0.2in}
\caption{Performances of SDF-Bayes varying the prior distribution}
\vskip 0.1in
\label{table:varying_prior2}
\begin{center}
	\begin{small}
		\setlength\tabcolsep{1.5pt}
		\begin{tabular}{c||c|c|c}
			\toprule
			Prior & \makecell{Safety vio.} & \makecell{Rec. errors} & \makecell{DLT observ.} \\
			\midrule
			Default & \textbf{0.023} & \textbf{0.203} & 0.296 \\
			\midrule
			HiVar & 0.008 & 0.212 & 0.288 \\
			\midrule
			NonInfo & 0.008 & 0.208 & \textbf{0.261} \\
			\bottomrule
		\end{tabular}
	\end{small}
\end{center}
\end{wraptable}

\subsection{Sensitivity of Prior Distribution}
\label{sec:sensitivity_prior}
We now provide the performances of SDF-Bayes varying the prior distribution of the parameters of the dose-toxicity model in Table \ref{table:varying_prior2}.
To show the sensitivity of the algorithms, we consider the non-informative prior distribution (NonInfo) (i.e., a uniform distribution) and the distributions with higher variances (HiVar) compared with the prior distributions in our default setting.
Specifically, for the non-informative prior, we use a uniform distribution that is truncated according to the domain of each parameter.
For the high variance prior, we use a normal distribution $\mathcal{N}(0,50)$ for $\theta_0$ and $\theta_3$ and use a gamma distribution with mean 1 and variance 10 (i.e., $\Gamma(0.1,0.1)$) for $\theta_1$ and $\theta_2$.

From the results, we can see that the performance of SDF-Bayes is similar regardless of the prior distribution of the parameters.
This implies that SDF-Bayes is robust to the prior distribution since the posterior distribution of the parameters is effectively constructed thanks to the cautious optimism.
Beside, it is worth noting that the prior distribution in the default setting is not a precise one as well \citep{riviere2014bayesian}.
Thus, the case with the non-informative prior distribution is an extreme case in clinical trials for drug combinations since the prior information on the parameters can be usually acquired from historical information, such as laboratory tests and clinical trials for a single drug, and characteristics of dose-combination models.
In conclusion, we can use SDF-Bayes in practice even in case with a lack of the prior information of drugs.

\begin{wraptable}{R}{0.47\linewidth}
	\vspace{-0.3in}
	\caption{Performances of SDF-Bayes varying $\psi_s$.}
	\label{table:safety_thres}
	\begin{center}
		\begin{small}
			\setlength\tabcolsep{1.5pt}
			\begin{tabular}{c||c|c|c|c}
				\toprule
				$\psi_s$ & \makecell{Safety viol.\\($\psi_s$)} & \makecell{Safety viol.\\($\xi+\epsilon_s$)} & \makecell{Rec. errors} & \makecell{DLT observ.} \\
				\midrule
				0.20 & 0.048 & 0.000 & 0.421 & 0.169 \\
				0.25 & 0.056 & 0.000 & 0.328 & 0.221 \\
				0.30 & 0.022 & 0.000 & 0.257 & 0.263 \\
				0.35 & 0.019 & 0.019 & 0.203 & 0.297 \\
				0.40 & 0.027 & 0.193 & 0.193 & 0.321 \\
				\bottomrule
			\end{tabular}
		\end{small}
	\end{center}
	\vspace{-0.1in}
\end{wraptable}

\subsection{Impact of Target Safety}
In clinical trials, the target toxicity threshold is a standard of safety, and thus, the target safety is usually defined by the target toxicity threshold $\xi$ ($\xi+\epsilon_s$ in our manuscript).
In SOTA Bayes, dose (de-)escalation is determined based on the target toxicity threshold.
Thus, its safety naturally focuses on the target toxicity threshold, and the target safety threshold cannot be arbitrarily determined.
On the other hand, SDF-Bayes can consider arbitrary target safety $\psi_s$ in its cautious principle by substituting the safety threshold, $\xi+\epsilon_s$, in residual $r(t)$ and the set of conservatice DCs $\mathcal{A}_c(t)$ in our manuscript, respectively, to any other target safety $\psi_s$.

Here, we show the impact of the safety threshold on SDF-Bayes.
In Table \ref{table:safety_thres}, the recommendation error rates, the safety violation rates with $\psi_s$, the safety violation rates with $\xi+\epsilon_s$, and the DLT observation rates are provided.
From the table, we can see that the safety violation rates with $\psi_s$ are similar regardless of the target safety, $\psi_s$.
This implies that SDF-Bayes effectively manages the risk of safety violation for any given target safety.
Consequently, the DLT observation rates increase according to the target safety.
This effective risk management in SDF-Bayes makes the target safety $\psi_s$ become a safety budget for the exploration of the toxicities of the DCs.
Thus, the larger $\psi_s$ allows SDF-Bayes to choose the DCs more optimistically during clinical trials.
Consequently, in the results, the recommendation error rates decreases as $\psi_s$ increases.
However, the target safety satisfying $\psi_s>\xi+\epsilon_s$ is hard to use in practice since the safety violation rate with the standard target safety in clinical trials (i.e., $\xi+\epsilon_s$) significantly increases.
On the other hand, the target safety satisfying $\psi_s<\xi+\epsilon_s$ can be used to make trials safer while sacrificing the error rates.

\subsection{Runtime and Scalability of SDF-Bayes}
We evaluated runtime of SDF-Bayes in our simulations with homogeneous group; on average, it took approximately 2.7 milliseconds to complete a one-round update. (The implementation employed MATLAB with Intel Core i7-8700 3.2GHz CPU but without parallel computing.) By contrast, in a clinical trial, it takes days or weeks (sometimes months) to enroll patients for each round. Besides, actual clinical trials  have limited numbers of patients and drug combinations. Even if the trial allowed for a large number of patients, testing  hundreds of drug combinations would not be realistic because of safety issues. Hence, the runtime of SDF-Bayes is  short enough that it can be used in any realistic trial.

\section{More Experiment Results with Heterogeneous Groups}

\subsection{Understanding of Behavior of Bayesian Algorithms Applied to Entire Population}

\begin{figure}[ht]
	\begin{center}
		\includegraphics[width=.55\columnwidth]{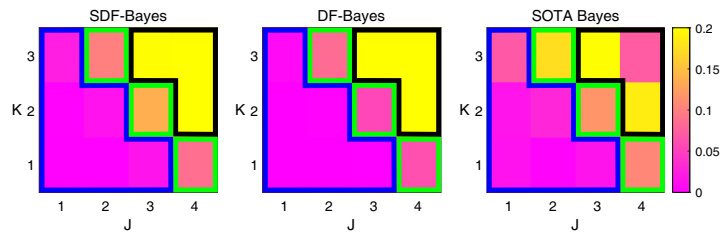}
	\vspace{-0.1in}
		\caption{Heatmaps for DC allocation ratios of the patients of group A with the Bayesian algorithms applied to entire population.}
		\label{fig:sel_heatmap_het_to_homo}
	\end{center}
	\vspace{-0.1in}
\end{figure}

Given the toxicities of group EP, the toxicities to group A are under-estimated because the toxicities to group B are very low.
As a result, SDF-Bayes and DF-Bayes choose the DCs close to the MTDs of the averaged synthetic model for the entire population (i.e., (2,4) and (3,3) as shown in Table \ref{table:toxicity_ep}).
Also, SOTA Bayes escalates dose above the MTDs of group A.
These imply that the algorithms applied to group EP frequently select overdosing DCs for the patients of group A; see Fig. \ref{fig:sel_heatmap_het_to_homo}.
Then, as shown in Table 4 in the manuscript, the safety violation rates of group A become excessively high.

\subsection{Expected Improvements and Patient Recruitment Ratios}

To understand the recruitment behavior of SDF-Bayes-AR, Fig. \ref{fig:recruit} plots the expected improvement (EI) probability for the most likely DC for each group and the patient recruitment ratio of each group as a function of the amount of prior information ($T_p$).
The EI for group B decreases as the amount of the prior information increases.
(Previous observations render later observations less useful.)  SDF-Bayes-UR recruits patients uniformly; SDF-Bayes-AR adaptively recruits patients in order to maximize the EI gain from the next patient.
\looseness=-1

\begin{figure}[ht]
	\begin{center}
		\includegraphics[width=.55\linewidth]{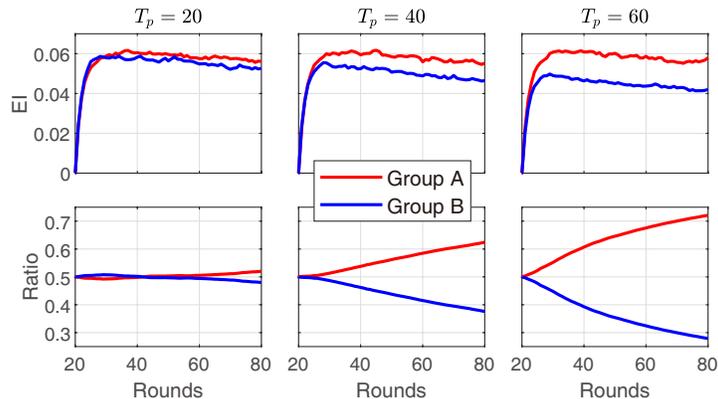}
		\caption{EI of the probability measures and patient recruitments ratio of groups varying the amount of the prior information.}
		\label{fig:recruit}
	\end{center}
\end{figure}

\vspace{-0.1in}
\subsection{Results in Section 5.2. with DLT Observation Rates}
Here, we provide the DLT observation rates of the results for heterogeneous groups.
Tables \ref{table:DLT_table3} and \ref{table:DLT_table4} provides the DLT observation rates aligned to the results in Tables 3 and 4 in our manuscript, respectively.
Table \ref{table:DLT_table3} shows  that the DLT observation rates of variants of SDF-Bayes are slightly higher than those of SOTA Bayes-UR due to the exploration-toxicity trade-off as in the single group case.
(It is shown that they satisfy the safety constraint in Table 3 in our manuscript.)
Table \ref{table:DLT_table4} shows that SDF-Bayes-AR outperforms SDF-Bayes-UR in terms of total DLT observation rates.

\begin{table}[h]
	\caption{DLT observation rates with heterogeneous groups (from Table 3 in the paper).}
	\label{table:DLT_table3}
	\begin{center}
		\begin{small}
			\setlength\tabcolsep{1.5pt}
			\begin{tabular}{c||c|c|c}
				\toprule
				Algorithms & ~~Total~~ & Group A & Group B \\
				\midrule
				SDF-Bayes-AR  & 0.263$\pm$.001 & 0.292$\pm$.001 & 0.236$\pm$.001 \\
				SDF-Bayes-UR  & 0.263$\pm$.001 & 0.297$\pm$.001 & 0.229$\pm$.001 \\
				DF-Bayes-UR   & 0.302$\pm$.001 & 0.352$\pm$.002 & 0.252$\pm$.002 \\
				SOTA Bayes-UR & \textbf{0.233$\pm$.001} & \textbf{0.270$\pm$.001} & \textbf{0.196$\pm$.001} \\
				\midrule
				SDF-Bayes-EP  & 0.278$\pm$.001 & 0.391$\pm$.002 & 0.166$\pm$.002 \\
				DF-Bayes-EP	  & 0.302$\pm$.001 & 0.427$\pm$.002 & 0.178$\pm$.002 \\
				SOTA Bayes-EP & 0.247$\pm$.001 & 0.352$\pm$.002 & 0.142$\pm$.001 \\
				\bottomrule
			\end{tabular}
		\end{small}
	\end{center}
\end{table}

\vspace{-0.1in}
\begin{table}[h]
	\caption{DLT observation rates varying the amount of the prior information (from Table 5 in the paper).}
	\label{table:DLT_table4}
	\begin{center}
		\begin{small}
			\setlength\tabcolsep{1.5pt}
			\begin{tabular}{c|c||c|c|c}
				\toprule
				\multicolumn{2}{c||}{} & $T_p=20$ & $T_p=40$ & $T_p=60$ \\
				\midrule
				\multirow{3}{*}{AR}
				& Group A & 0.282$\pm$.001 & 0.261$\pm$.001 & 0.250$\pm$.001 \\
				& Group B & 0.243$\pm$.002 & 0.259$\pm$.002 & 0.269$\pm$.003 \\
				& Entire  & 0.262$\pm$.001 & 0.255$\pm$.001 & 0.248$\pm$.001 \\
				\midrule
				\multirow{3}{*}{UR}
				& Group A & 0.296$\pm$.001 & 0.296$\pm$.001 & 0.296$\pm$.001 \\
				& Group B & 0.238$\pm$.002 & 0.248$\pm$.002 & 0.255$\pm$.002 \\
				& Entire  & 0.267$\pm$.001 & 0.272$\pm$.001 & 0.275$\pm$.001 \\
				\midrule
				\bottomrule
			\end{tabular}
		\end{small}
	\end{center}
\end{table}

%
%
%


\end{document}